\RequirePackage{fix-cm}
\documentclass[12pt]{article}
\PassOptionsToPackage{table,dvipsnames}{xcolor}

\usepackage[left=2.5cm,
right=2.5cm,
top=2.3cm,
bottom=2.3cm,
headheight=20pt,
headsep=10pt,
footskip=25pt,
letterpaper]{geometry}
\usepackage[utf8]{inputenc}
\usepackage[T1]{fontenc}
\usepackage[english]{babel}
\usepackage{amsmath,amsfonts,amssymb,amsthm,thmtools}
\usepackage{graphicx}
\usepackage{hyperref}
\usepackage{fancyhdr}
\usepackage[normalem]{ulem}
\usepackage{graphicx}
\usepackage{stfloats}
\usepackage{wrapfig}
\usepackage{epstopdf}
\usepackage{cleveref}
\usepackage{subfloat}
\usepackage{subcaption}
\usepackage{xspace}
\usepackage{enumitem}
\usepackage{listings}
\usepackage{titlesec}
\usepackage{etoolbox}
\usepackage{setspace}
\usepackage{changepage}
\usepackage{etoolbox}
\usepackage{multirow}
\usepackage{booktabs}
\usepackage{tabularx}
\usepackage{wrapfig}
\usepackage{svg}
\usepackage[percent]{overpic}
\usepackage[round]{natbib}
\usepackage[colorinlistoftodos, shadow,color=blue!30!white
]{todonotes}
\usepackage{xpatch}
\usepackage{siunitx}

\fancypagestyle{first}{\fancyfoot[R]{\small\thepage}}

\setlist[itemize]{leftmargin=1em,itemsep=0ex,topsep=0ex}
\titlespacing*{\paragraph}{0pt}{0ex plus .1ex}{1ex}
\titlespacing*{\section}{0ex}{2.3ex plus .3ex minus .0ex}{.6ex plus .3ex minus .2ex}
\titlespacing*{\subsection}{0ex}{1.5ex plus .3ex minus .5ex}{.4ex plus .2ex minus .1ex}
\titlespacing*{\subsubsection}{0ex}{1.2ex plus .3ex minus .3ex}{.3ex plus .2ex minus .2ex}

\xapptocmd\normalsize{%
\abovedisplayskip=.8em plus .2em minus .2em
\belowdisplayskip=.6em plus .1em minus .1em
\abovedisplayshortskip=.8em plus .2em minus .2em
\belowdisplayshortskip=.6em plus .1em minus .1em
}{}{}

\setcitestyle{numbers}
\renewcommand{\cite}[1]{\citep{#1}}

\definecolor{mydarkblue}{rgb}{0.0,0.15,0.7}
\hypersetup{%
colorlinks=true,
linkcolor=mydarkblue,
citecolor=mydarkblue,
filecolor=mydarkblue,
urlcolor=mydarkblue}

\makeatletter

  \renewcommand{\maketitle}{%
    \begingroup
      {\centering\LARGE\@title\par}%
      \vskip 1em
      \centering
      \begin{tabular}[t]{@{}c@{}}\strut\@author\strut\end{tabular}%
      \vskip 0.3in minus 0.1in
    \endgroup
  }
\makeatother

\PassOptionsToPackage{table,dvipsnames}{xcolor}
\usepackage{amsmath,amssymb}

\usepackage[T1]{fontenc}
\definecolor{gainred}{RGB}{190,45,45}

\newcommand{\badgain}[1]{%
  \hspace{2pt}{\scriptsize\textcolor{gainred}{(#1)}}%
}
\usepackage[utf8]{inputenc}
\usepackage{graphicx}
\usepackage{wrapfig}
\usepackage{microtype,placeins}
\usepackage{algorithm,algpseudocode}
\usepackage{hyperref}
\usepackage{url}
\hypersetup{
    colorlinks = true,
    citecolor  = OliveGreen,
    linkcolor  = Blue,
    urlcolor   = Plum
}
\DeclareMathOperator*{\argmax}{arg\,max}

\DeclareMathOperator{\Std}{Std}

\newcommand{\method}{\textsc{DRSR}}

\RequirePackage[table]{xcolor}
\RequirePackage{booktabs,array,tabularx,multirow,makecell,longtable}
\RequirePackage{caption}
\newcolumntype{Y}{>{\centering\arraybackslash}X}
\newcolumntype{L}{>{\raggedright\arraybackslash}X}
\RequirePackage{needspace}
\definecolor{NoteText}{gray}{0.25}
\newcommand{\best}[1]{\textbf{#1}}
\newcommand{\second}[1]{\underline{#1}}
\newcommand{\tableformat}{%
  \fontsize{9.6}{11.4}\selectfont
  \setlength{\tabcolsep}{5pt}%
  \renewcommand{\arraystretch}{1.10}%
  \setlength{\aboverulesep}{1.5pt}%
  \setlength{\belowrulesep}{1.5pt}%
  \setlength{\arrayrulewidth}{0.35pt}}

\newcommand{\doublerule}{%
  \specialrule{0.65pt}{1.5pt}{1.0pt}%
  \specialrule{0.35pt}{0pt}{1.5pt}}
\newcommand{\tablenote}[1]{%
  \par\vspace{3pt}{\fontsize{8.6}{10.2}\selectfont\color{NoteText}\raggedright #1\par}}
\title{DRSR: Learning Set-Level Deletion Risk for Efficient Long-Horizon Agents}
\author{%
  \mbox{Mingxuan Wang\textsuperscript{1}}\quad \mbox{Bo Wang\textsuperscript{1}}\quad \mbox{Fei Luo\textsuperscript{1}}\quad \mbox{Guorun Yao\textsuperscript{1}}\quad \mbox{Chao Ning\textsuperscript{1}}\\[2pt]
  \mbox{Yinglong Guo\textsuperscript{1}}\quad \mbox{Hongyue Chen\textsuperscript{1}}\quad \mbox{Yanbiao Ma\textsuperscript{2,*}}\quad \mbox{Jungong Han\textsuperscript{3,*}}\\[4pt]
  \textsuperscript{1}TierFlow Team\\
  \textsuperscript{2}Gaoling School of Artificial Intelligence, Renmin University of China\\
  \textsuperscript{3}Tsinghua University\\[3pt]
  \textsuperscript{*}Corresponding authors.\quad \href{mailto:ybma1998@ruc.edu.cn}{\texttt{ybma1998@ruc.edu.cn}}%
}
\hypersetup{pdfauthor={Mingxuan Wang, Bo Wang, Fei Luo, Guorun Yao, Chao Ning, Yinglong Guo, Hongyue Chen, Yanbiao Ma, Jungong Han}}

\date{}
\hypersetup{colorlinks=true,linkcolor=mydarkblue,citecolor=mydarkblue,urlcolor=mydarkblue}
\usepackage{team-template}
\renewcommand{\TeamPaperID}{DRSR / LONG-HORIZON AGENTS}
\renewcommand{\TeamShortTitle}{DRSR}
\begin{document}
\pagestyle{fancy}
\maketitle
\thispagestyle{first}

\begin{abstract}
Long-horizon language-model agents accumulate reasoning traces, tool exchanges, and observations whose relevance changes with the current decision. Existing compression strategies often score historical units independently, but the safety of deleting several units is generally not determined by their singleton scores: redundant evidence, accumulated small effects, and the information that remains after deletion all matter. We introduce \textbf{Direct Relational Set-Risk Pruning (DRSR)}, which formulates agent-history compression as risk-constrained selection over deletion sets. Offline, DRSR constructs exact counterfactual supervision by jointly deleting protocol-valid history Blocks and measuring the change in teacher-forced likelihood of the same recorded next output. A lightweight scorer then predicts set-level harm from online-visible relations between candidate history and the current pre-action state, together with deleted--retained and pairwise set structure. At deployment, DRSR evaluates a small set of structurally valid deletion candidates with the lightweight scorer and removes the largest feasible set under recency, protocol, budget, and learned-risk constraints, abstaining when no set is sufficiently safe. On WorkBuddyBench Full260, DRSR increases mean reward from 0.699 to 0.802 while reducing total model tokens by 20.820\%. On the fixed Eval40 comparison, it obtains 0.794 reward at 1.211M tokens per task, using 35.850\% fewer tokens than the uncompressed agent. Mechanistic analyses and ablations further show that decision-conditioned relations, retained-context information, pair interactions, and abstention each contribute to reliable pruning.
\end{abstract}

\section{Introduction}
\label{sec:introduction}

Agents solve long horizon tasks by interleaving reasoning with tool use and
incorporating observations into later decisions \citep{yao2022react}. During
execution, their context accumulates task constraints, explored alternatives,
intermediate conclusions, repeated observations, and completed tool
interactions. Keeping the full history avoids information loss, but it also
requires the model to repeatedly process an increasingly long context, much of
which may no longer matter to the current decision at later stages of execution.
As context grows, separating useful evidence from obsolete or redundant content
becomes more difficult \citep{liu2024lost,shi2023large}. Existing prompt compression
and memory management methods reduce active context
\citep{jiang2023llmlingua,pan2024llmlingua,packer2023memgpt}, but safe
history pruning requires identifying which past information is still needed
at the current agent state.

Historical value is inherently state dependent. The importance of a past
record is not a fixed property of the record itself, but changes as the agent
progresses through the task, motivating recent work on adaptive context
management for long-horizon agents
\citep{li2026sculptor,zhang2026memory,kang2025acon}.
An error trace may be essential while a failure
is being diagnosed, yet become redundant once the cause has been identified
and the relevant conclusion has been recorded elsewhere. In contrast, an
early task constraint may remain important until the task is completed.
Historical information therefore cannot be judged reliably from age,
position, or a fixed context window alone. Its value depends on the current
decision state and on what information would remain after pruning.

Safe pruning is also a set level problem. Two historical Blocks may each
appear removable in isolation because either one preserves the same fact,
while removing both may eliminate the only remaining evidence. More generally,
removal-based attribution and interaction methods show that the effect of a
set need not be recoverable from independently measured singleton effects
\citep{covert2021explaining,grabisch1999axiomatic,
sundararajan2020shapley,tsai2023faith}.
Even without such interaction, several small deletion effects may accumulate
into substantial harm. Independent Block scores therefore cannot reliably
recover the risk of the set that is actually removed.

The main difficulty is that true deletion harm is unavailable when the online
decision must be made. The effect of removing history is defined with respect
to the agent's actual next output, which has not yet occurred. This distinction
is related to learning with privileged information and distillation, where
training may exploit information unavailable at deployment
\citep{vapnik2009new,hinton2015distilling,lopez2015unifying}.
Repeatedly querying the language model for every candidate deletion is also
unsuitable, because different queries may produce different continuations and
would make pruning itself expensive. Related learned eviction methods likewise
transfer future or privileged signals into lightweight online selectors
\citep{dong2026foresightkv}.
Completed trajectories make this quantity available for supervision. Once the
actual next output is known, we can keep it fixed, jointly remove a candidate
set, and measure how its likelihood changes under the edited context. This
converts an online unavailable quantity into an offline training signal. A
lightweight predictor can then learn deletion risk from information available
before the next action.

\begin{figure}[t]
    \centering
    \includegraphics[width=\linewidth]{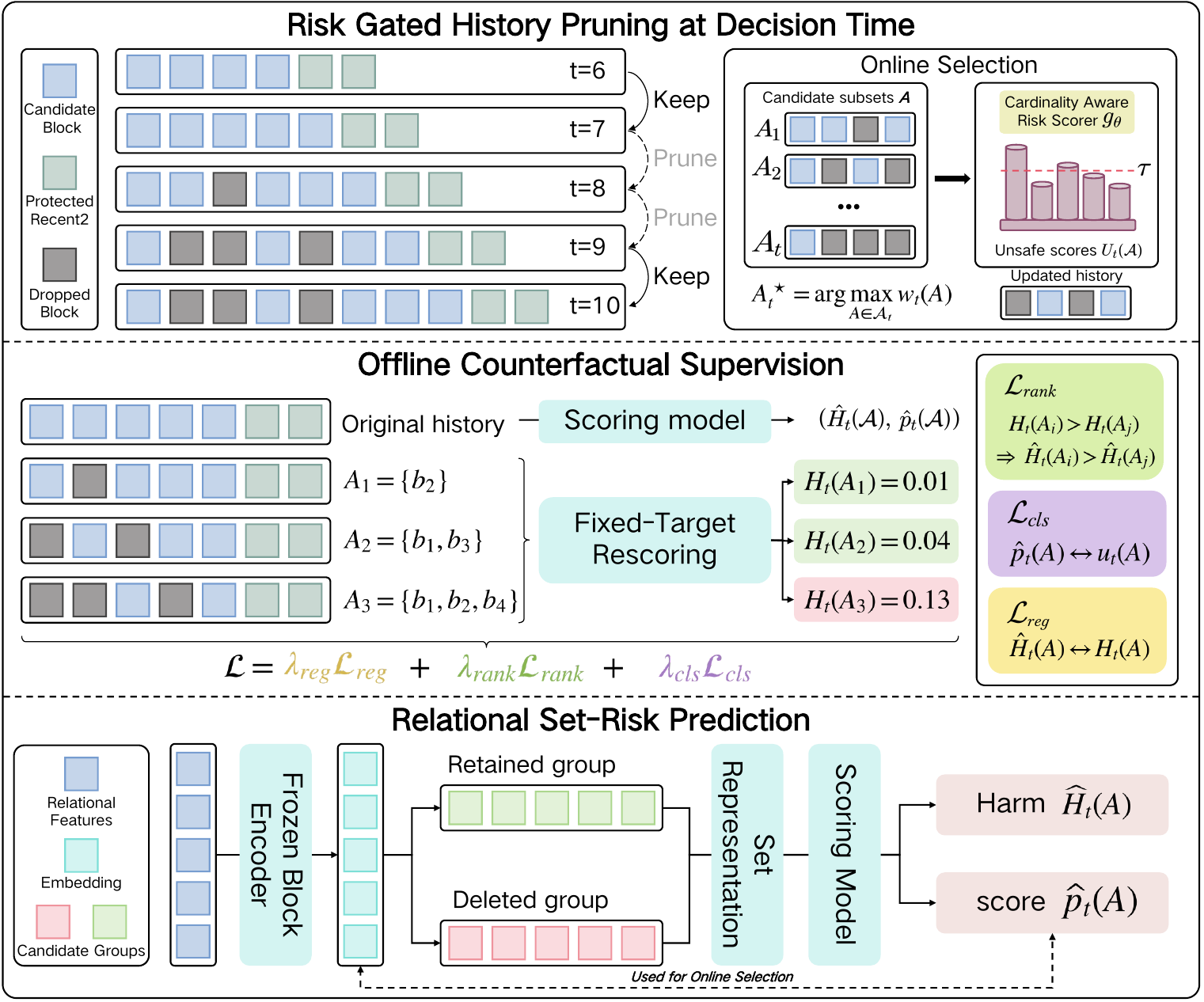}
    \caption{\textbf{Overview of DRSR.}
    Offline, DRSR constructs exact set-level counterfactual supervision by
    jointly deleting protocol-valid history Blocks and rescoring the same
    recorded next output. A relational set-risk predictor models both the
    proposed removal set and the information that remains, and outputs
    predicted harm and an unsafe score. Online, the learned score gates
    candidate deletion sets, after which DRSR selects the feasible set with
    the largest direct token saving and abstains when no set is sufficiently
    safe.}
    \label{fig:overview}
\end{figure}

Based on this idea, we introduce \textbf{Direct Relational Set Risk Pruning
(\method{})}. \method{} represents historical value through relations between
candidate history and the current decision state, while modeling both the
information proposed for removal and the information that remains. Instead of
assigning each Block a fixed importance score, it directly predicts the risk
of a candidate deletion set. During deployment, candidate sets are evaluated
in a batch by a lightweight scorer, with no future output and no additional
language model evaluation for each deletion. Pruning is allowed only when the
candidate satisfies protocol constraints, recent history protection,
deletion limits, and a learned risk gate. If no candidate is sufficiently
safe, the original request is retained, following the broader idea of
selective prediction with abstention
\citep{chow1970optimum,geifman2019selectivenet}.
Figure~\ref{fig:overview}
summarizes the offline supervision, relational set-risk prediction, and
online risk-gated selection pipeline. Our main contributions are:

{
\setlength{\leftmargini}{1.2em}
\setlength{\itemsep}{3pt}

\begin{itemize}

    \item \textbf{State conditioned set level supervision.}
    We construct deletion labels from completed trajectories by jointly
    removing candidate Blocks and rescoring the same recorded next output.
    Each label therefore measures the proposed deletion set directly rather
    than combining independently evaluated singleton scores.

    \item \textbf{Relational set risk prediction.}
    We introduce a lightweight scorer that jointly models the current
    decision state, candidate history, deleted information, retained
    information, and interactions within the deletion set. The representation
    backbone and task model remain frozen.

    \item \textbf{Risk constrained online pruning with abstention.}
    We use predicted risk as a feasibility condition and select the feasible
    deletion set with the largest direct token saving. If no candidate passes
    the risk gate, the policy abstains from pruning.

\end{itemize}
}

\section{Related Work}
\label{sec:related_work}

\paragraph{Agent memory and context management.}
Long-horizon agents increasingly manage their own active context. MemGPT
organizes information across memory tiers \citep{packer2023memgpt}, while
Sculptor, Memory-as-Action, Git-Context-Controller, context folding, and
AgentFold expose or learn explicit history editing operations
\citep{li2026sculptor,zhang2026memory,wu2025git,sun2025scaling,ye2025agentfold}.
ACON optimizes compression policies from trajectory outcomes
\citep{kang2025acon}, whereas LLMLingua and LLMLingua-2 compress prompts under
a token budget \citep{jiang2023llmlingua,pan2024llmlingua}. These approaches
are commonly evaluated end to end on interactive benchmarks
\citep{trivedi2024appworld,jimenez2024swe}. DRSR instead supervises each
protocol-valid deletion set by its measured effect on a fixed recorded next
output, without replacement summaries or retrieval stores.

\paragraph{Structured pruning and KV-cache eviction.}
Reasoning traces can be shortened by generating less per step, skipping
low-importance tokens, or removing low-information steps
\citep{xu2025chain,xia2025tokenskip,li2026making}. Related inference-time
methods evict KV entries using attention, recency, adaptive budgets, or learned
future contribution \citep{zhang2023h2o,xiao2024efficient,li2024snapkv,
feng2026ada,dong2026foresightkv}. DRSR differs in operating on serialized
agent-history Blocks and directly predicting the risk of a jointly deleted
set rather than assigning independent importance scores to individual units.

\paragraph{Removal-based measures and interactions.}
Removal-based explanation studies how behavior changes when subsets of
features are removed \citep{covert2021explaining}. The interaction residual
used by DRSR is closely related to discrete Shapley interaction terms
\citep{grabisch1999axiomatic,lundberg2018consistent,
sundararajan2020shapley,tsai2023faith,fumagalli2023shap}. However, DRSR does
not seek symmetric attribution over all historical Blocks; it measures and
predicts the harm of the specific deletion sets permitted by the agent
protocol.

\paragraph{Privileged supervision and selective prediction.}
Using training-only information to learn a deployable predictor is related to
learning with privileged information and knowledge distillation
\citep{vapnik2009new,hinton2015distilling,lopez2015unifying}. DRSR also
abstains when no candidate passes its learned risk gate, connecting it to
reject-option and risk-controlled prediction
\citep{chow1970optimum,el2010foundations,geifman2019selectivenet,
bates2021distribution,angelopoulos2024conformal,mohri2024language}.
Semantic-uncertainty methods provide a complementary line of work
\citep{kuhn2023semantic,farquhar2024detecting,nikitin2024kernel}; DRSR does
not estimate semantic uncertainty or provide a conformal guarantee.

\section{Problem Formulation}
\label{sec:methodology}

\subsection{Structured History and Deletion Actions}
\label{sec:history}

At decision time $t$, let $P_t$ denote the protected system instructions
and task specification, and let $S_1,\ldots,S_{t-1}$ denote the completed
interaction Steps. We write
\begin{equation}
C_t
=
P_t\oplus S_1\oplus\cdots\oplus S_{t-1},
\qquad
S_i
=
B_{i,0}\oplus\cdots\oplus B_{i,m_i},
\label{eq:history}
\end{equation}
where $\oplus$ denotes serialization in the original order. A Block is
either the complete assistant content associated with a Step
(\texttt{assistant\_state}) or an atomic tool exchange that contains a
tool call together with its matched response (\texttt{tool\_exchange}).
Pruning never separates a tool call from its response, and every rewritten
request must preserve a valid conversation and tool protocol.

Let $E_t=\{B_1,\ldots,B_n\}$ denote the eligible candidate pool and let
$A\subseteq E_t$ denote a proposed deletion set. The operator $D_A(C_t)$
jointly removes all Blocks in $A$ and serializes the remaining context in
its original order. We define the direct token saving as
\begin{equation}
w_t(A)
=
\operatorname{Tok}(C_t)
-
\operatorname{Tok}(D_A(C_t)).
\label{eq:saving}
\end{equation}
The saving is measured on the actual rewritten request. We therefore do
not assume that the saving of a deletion set equals the sum of token counts
measured for its individual Blocks.

\subsection{Counterfactual Deletion Harm}
\label{sec:harm}

For an external trajectory, let $y_t$ denote the recorded next assistant
output of length $m$. The target excludes later environment responses and
is held fixed when comparing the original context with a deleted context.
Given a frozen scoring language model $p_\omega$, we define
\begin{equation}
\mathcal L_\omega(y_t\mid C)
=
-\frac{1}{m}\sum_{k=1}^{m}
\log p_\omega\!\left(y_{t,k}\mid C,y_{t,<k}\right),
\qquad
H_t(A)
=
\mathcal L_\omega(y_t\mid D_A(C_t))
-
\mathcal L_\omega(y_t\mid C_t).
\label{eq:nll}
\end{equation}
The quantity $H_t(A)$ measures how much harder it becomes to reproduce the
same recorded decision after deleting $A$. Holding $y_t$ fixed makes the
comparison depend on the deletion rather than on a newly sampled
continuation. Normalization by output length places decisions of different
lengths on a common scale. Since all candidate sets at a state are measured
from the same intact context, the original context score can be reused
when constructing the corresponding labels.

We define an offline unsafe event by
\begin{equation}
u_t(A)=\mathbf 1[H_t(A)>\delta_H],
\label{eq:label}
\end{equation}
where $\delta_H$ is a fixed harm tolerance. Importantly, deletion harm is
indexed by the current decision state. The same historical Block may be
removable at one stage of execution and necessary at another. Historical
value is therefore treated as state dependent rather than as a fixed
property of the Block itself. The value used for $\delta_H$ and its
interpretation are reported in Appendix~\ref{app:setup}.

\subsection{Why Singleton Scores Do Not Determine Set Risk}
\label{sec:joint_formulation}

For a single Block, define $H_{i,t}=H_t(\{B_i\})$. For $B_i\notin A$, its
conditional effect after deleting $A$ is
\[
H_t(B_i\mid A)=H_t(A\cup\{B_i\})-H_t(A).
\]
If $A_k=\{B_{i_1},\ldots,B_{i_k}\}$ and all intermediate rewrites are
valid, then
\begin{equation}
H_t(A)
=
\sum_{k=1}^{|A|}H_t(B_{i_k}\mid A_{k-1}),
\qquad
A_0=\varnothing.
\label{eq:telescoping}
\end{equation}
Thus, joint harm is a sum of effects evaluated under changing retained
contexts. It is not generally the sum of singleton effects measured in the
original context. Equation~\eqref{eq:telescoping} holds for every deletion
order, while the final quantity $H_t(A)$ is independent of that order. The
conditional terms change because each deletion changes what information
remains.

To measure this difference directly, we define
\begin{equation}
I_t(A)
=
H_t(A)-\sum_{B_i\in A}H_{i,t},
\qquad
J_{ij,t}
=
H_t(\{B_i,B_j\})-H_{i,t}-H_{j,t}.
\label{eq:interaction}
\end{equation}
When $I_t(A)>0$, the actual joint deletion is more harmful than singleton
addition suggests. This can occur when two Blocks contain redundant
evidence: either one appears safe to remove while the other remains, but
removing both eliminates the last copy of that evidence. When $I_t(A)<0$,
singleton addition overestimates joint harm because it counts part of the
same effect more than once. Only $I_t(A)=0$ permits exact additive
reconstruction from singleton harms.

Interaction is not the only limitation of singleton scoring. Even when
$I_t(A)=0$, several individually acceptable harms can accumulate beyond the
allowed tolerance. DRSR therefore treats history pruning as a bounded set
selection problem. Direct set-level prediction addresses interactions among
deleted Blocks and the effect of what remains, while structural constraints
limit excessive deletion. These two failure modes motivate predicting risk
for the proposed set rather than deriving it from independently scored
historical units.

\subsection{Risk-Constrained Deletion Objective}
\label{sec:information}
\label{sec:oracle_objective}

Let $\mathcal A_t$ denote the deletion sets that satisfy the structural
requirements of the agent protocol. If exact deletion harm were observable
at decision time, an ideal pruning policy would solve
\begin{equation}
A_t^{\mathrm{oracle}}
\in
\argmax_{A\in\mathcal A_t}w_t(A)
\qquad
\text{subject to}
\qquad
H_t(A)\le\delta_H.
\label{eq:oracle_selection}
\end{equation}
This objective separates compression benefit from deletion risk. The
quantity $w_t(A)$ measures how much context is removed, while $H_t(A)$
determines whether the proposed edit remains within the allowed harm
tolerance.

The exact constraint cannot be evaluated online. Computing $H_t(A)$ requires
the realized next output $y_t$, which does not yet exist when the pruning
decision is made. Evaluating candidate deletions through separate model
generations would also fail to recover the fixed-target quantity in
Eq.~\eqref{eq:nll}, because different contexts can produce different
continuations. Completed trajectories make exact supervision possible
offline: after $y_t$ is observed, candidate sets can be deleted jointly and
the same recorded output can be rescored.

DRSR learns an online scorer from these offline measurements. At deployment,
the scorer estimates deletion risk using only the current causal context and
the proposed deletion set during each online pruning decision. It therefore
replaces the unavailable oracle constraint, while direct token saving remains
the quantity optimized by the pruning policy.

\section{Direct Relational Set-Risk Pruning}
\label{sec:drsr}

\subsection{Candidate Construction and Relational Features}
\label{sec:features}

At each eligible decision, DRSR protects recent complete Steps and constructs
a small candidate pool from older protocol-valid history. Candidates are
selected across the trajectory using chronological stratification, with
preference given to edits that offer meaningful direct token saving. The
candidate rule is fixed before any deletion outcome is observed, so the
prediction problem is not simplified using label information. Exact
candidate counts, protection rules, alignment checks, and context limits are
reported in Appendix~\ref{app:candidate_features}.

A frozen representation backbone encodes the intact causal prefix. DRSR does
not assign an intrinsic importance score to a historical Block. Instead,
each candidate $B_i$ is represented through its relation to the current
decision state. Let $\Omega_i$ denote the token span of $B_i$. The feature
vector $x_{i,t}$ combines structural controls such as age, position, type,
length, lexical overlap, tool overlap, and redundancy with attention and
hidden-state relations to the current pre-action state.

For a selected attention layer $\ell$ and head $h$, we define attention mass
and density as
\begin{equation}
a_{i,t}^{\ell,h}
=
\sum_{k\in\Omega_i}\alpha_{t,k}^{\ell,h},
\qquad
d_{i,t}^{\ell,h}
=
\frac{a_{i,t}^{\ell,h}}{|\Omega_i|}.
\label{eq:attention}
\end{equation}
Attention mass measures how much attention from the current pre-action
position falls on the Block, while density reduces the direct effect of
Block length.

Hidden-state relations provide a complementary view of the same Block to
state relation. DRSR compares the mean hidden representation of the Block
with the hidden representation at the current pre-action position using
cosine similarity, normalized distance, relative norm, and boundary-token
relations across selected layers. These features describe whether the
historical content aligns with the current decision in representation space,
rather than describing the Block in isolation. Fixed, label-independent
projections retain additional directional information that a single
similarity value cannot capture. Exact formulas, layers, dimensions, and
projection details are given in Appendix~\ref{app:features}.

Attention, hidden-state relations, and projected semantic features are inputs
to the learned risk model, not deletion rules by themselves. Low attention
does not imply that a Block is safe to remove, and high similarity does not
imply that it must be retained. The final prediction is supervised by
measured set-level deletion harm and also depends on the information that
survives the proposed edit.

\subsection{Set-Risk Representation and Learning}
\label{sec:set_representation}
\label{sec:training}

For a proposed deletion $A\subseteq E_t$, let $D=A$ denote the candidates to
be removed and let $R=E_t\setminus A$ denote the candidates that remain. DRSR
represents the proposed edit as a whole rather than assigning each Block a
fixed deletion score:
\begin{equation}
r_t(A)
=
\operatorname{SetEnc}_{\theta}
\left(
\{x_{i,t}:B_i\in D\},
\{x_{i,t}:B_i\in R\},
b_t(A)
\right).
\label{eq:setrep}
\end{equation}
The encoder represents four sources of information: what is removed, what
remains, relations among jointly deleted candidates, and relations between
deleted and retained evidence. Deletion risk therefore depends not only on
the content being removed but also on whether equivalent or supporting
information survives elsewhere in the context.

A lightweight pairwise summary gives the scorer direct information about
relations within the deleted set. This provides an interaction-aware bias
without assuming that final risk is a sum of pairwise terms. The exact
pooling operators, deleted and retained summaries, and budget features are
given in Appendix~\ref{app:exact_setrep}.

A lightweight cardinality-aware scorer maps the set representation to
predicted harm and an unsafe score for each candidate deletion set:
\begin{equation}
\bigl(\widehat H_t(A),\widehat p_t(A)\bigr)
=
g_{\theta,|A|}\!\left(r_t(A)\right).
\label{eq:setrisk}
\end{equation}
The two outputs serve complementary roles. Predicted harm preserves the
continuous ordering of deletion effects, while the unsafe score directly
supports the thresholded online decision. Training combines harm regression,
ranking among deletion sets, and unsafe-set classification. The unsafe score
is used for decision making and is not interpreted as a probability of task
failure. Exact model dimensions, loss terms, and optimization settings are
reported in Appendices~\ref{app:features} and~\ref{app:optimization}.

\subsection{Risk-Gated Selection and Abstention}
\label{sec:selection}

At deployment, the learned risk score is used as a feasibility condition
rather than as a reward for compression. Candidate subsets are evaluated
only by the lightweight set-risk scorer. DRSR does not delete each candidate
set and rerun the language model to estimate its harm online. Protocol
validity, recent-history protection, deletion size, and the removal budget
are enforced directly, while the learned score estimates whether an
otherwise valid edit is sufficiently low risk.

Let $\mathcal A_t$ denote the sets that satisfy the structural constraints,
and let $U_t(A)$ denote the aggregate unsafe score produced by the scorer
ensemble. DRSR selects
\begin{equation}
A_t^\star
\in
\argmax_{A\in\mathcal A_t}w_t(A)
\qquad
\text{subject to}
\qquad
U_t(A)\le\tau.
\label{eq:selection_objective}
\end{equation}
This is the online counterpart of Eq.~\eqref{eq:oracle_selection}. The
unavailable exact harm constraint is replaced by a learned score that can be
computed before the next action under the current causal context, while direct
token saving remains the objective. If no candidate set passes the risk gate,
DRSR abstains and returns the original request unchanged.

The candidate space is deliberately small, so all admissible subsets can be
scored together by the lightweight network. Online execution performs no
candidate-specific counterfactual deletion forward and no additional
language model query for each proposed deletion. The exact candidate limit,
removal budget, ensemble rule, validation-selected threshold, tie breaking,
and computational cost are reported in Appendix~\ref{app:deployment_details}
and Algorithm~\ref{alg:drsr}.

\section{Experiments}
\label{sec:experiments}

We evaluate DRSR from both offline prediction and end-to-end agent
performance for a comprehensive system-level assessment. The offline scorer is trained on 566 external trajectories
containing 1,698 decision states and approximately 24,122 jointly measured
deletion-set labels, split 80:10:10 by source trajectory and disjoint from
WorkBuddyBench. End-to-end evaluation uses WorkBuddyBench Full260
(80 Code, 50 Office, 60 Security, and 70 Web tasks) and a fixed Eval40
subset with ten tasks per domain. Offline analyses report meaningful-gap
pair accuracy, within-state Spearman correlation, unsafe-event AUROC, and
Brier score; agent evaluation reports task reward and input-plus-output
model tokens. Full260 aggregates are task-count weighted, whereas Eval40
uses the equal-domain mean. Additional dataset, token-accounting,
task-manifest, and information-boundary details are provided in
Appendices~\ref{app:setup}, \ref{app:optimization},
and~\ref{app:deployment_details}.

\subsection{Mechanistic Analysis of History and Set Risk}
\label{sec:mechanistic_analysis}
\label{sec:history_results}
\label{sec:joint_results}

\paragraph{State-conditioned historical relevance.}
The granularity analysis in Appendix~\ref{app:extra_figures} shows that
the median within-state harm range is 0.312 at Step level but only 0.028
at Block level, and 71.400\% of unsafe parent Steps contain at least one
safe constituent Block. This motivates Block-level actions. More
importantly, historical value is decision-conditioned. Using the matched
pre-action state yields 70.920\% meaningful-gap pair accuracy, compared
with 62.020\% under the mismatched-state control, an 8.900-point gap with
the candidate Block held fixed. The full-context relation reaches
66.390\%, while the best combined relation reaches 72.130\%, indicating
that relating a candidate to the correct current decision state is more
informative than adding context indiscriminately. Complete granularity,
layerwise, and harm-margin results are reported in
Appendix~\ref{app:extra_figures}.

\paragraph{Joint-deletion effects.}

\begin{wraptable}{r}{0.31\linewidth}
    \vspace{-0.95\baselineskip}
    \centering

    \captionsetup{
        justification=centering,
        singlelinecheck=true,
        skip=3pt
    }

    \caption{Singleton aggregation.}
    \label{tab:singleton_aggregation}

    \fontsize{8.7}{10.0}\selectfont
    \renewcommand{\arraystretch}{1.05}

    \begin{tabular}{@{}lr@{}}
    \toprule[0.8pt]
    \textbf{Rule} & \textbf{Pair Acc.} \\
    \midrule
    Sum
    & \best{58.94\%} \\

    Mean
    & 56.81\% \badgain{-3.61\%} \\

    Maximum
    & 46.17\% \badgain{-21.67\%} \\

    Minimum
    & 14.89\% \badgain{-74.74\%} \\

    \bottomrule[0.8pt]
    \end{tabular}

    \vspace{-0.15\baselineskip}
\end{wraptable}
The central structural question is whether singleton importance can
recover the risk of deleting a \emph{set}. We first test the most direct
alternative to set-level prediction: singleton harms are aggregated and
used to rank candidate deletion sets.
Table~\ref{tab:singleton_aggregation} shows that this approximation is
unreliable. Even the strongest simple rule, summation, reaches only
58.940\% meaningful-gap pair accuracy. Mean aggregation performs
similarly, while maximum and minimum aggregation deteriorate further.
Thus, set risk cannot be reliably recovered from independently evaluated
singleton effects.

\Needspace{190pt}
\begin{wrapfigure}{r}{0.53\linewidth}
    \vspace{-0.75\baselineskip}
    \centering

    \includegraphics[width=\linewidth]{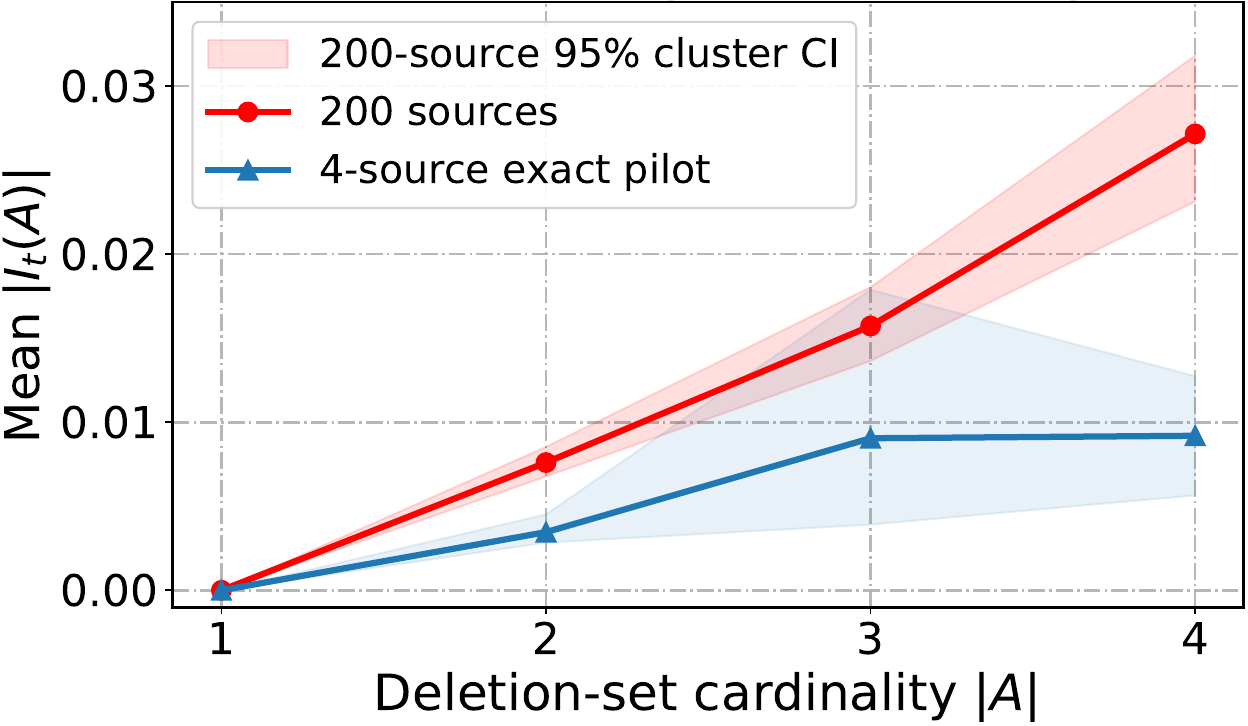}
    \vspace{-1.05\baselineskip}

    \captionsetup{
        justification=centering,
        singlelinecheck=true
    }

    \caption{Interaction grows with cardinality.}
    \label{fig:cardinality_main}

    \vspace{-1.05\baselineskip}
\end{wrapfigure}

The failure of singleton aggregation is consistent with the measured
interaction residual. Figure~\ref{fig:cardinality_main} examines this
effect over the 200-source confirmatory range. Mean absolute
non-compositionality increases from 0.000 for singleton deletion to
0.008, 0.016, and 0.027 for deletion sets of size two, three, and four,
respectively.

The discrepancy between independent singleton effects and actual joint
deletion therefore becomes more pronounced as several Blocks are removed
together. At three Blocks, the mean interaction magnitude is already
close to the $\delta_H=0.020$ harm-label threshold, while at four Blocks
it exceeds that threshold. This makes additive singleton scoring
increasingly unreliable in the multi-Block regime targeted by DRSR.

The complete analysis, including the exploratory extension to larger
sets and the per-domain pilot, is reported in
Appendix~\ref{app:extra_figures}.

We next ask whether explicit interaction information can explain the
discrepancy between singleton and joint deletion effects.

\begin{wraptable}{r}{0.35\linewidth}
    \vspace{-1\baselineskip}
    \centering

    \caption{Joint-harm reconstruction.}
    \label{tab:pair_reconstruction}

    \fontsize{8.7}{10.1}\selectfont
    \setlength{\tabcolsep}{3.6pt}
    \renewcommand{\arraystretch}{1.05}

    \begin{tabular*}{\linewidth}{@{\extracolsep{\fill}}lr@{}}
    \toprule[0.8pt]
    \textbf{Method} & \textbf{MAE} \\
    \midrule
    Singleton additive
        & 0.018 \\
    + exact pair interaction
        & \best{0.008} \\
    \bottomrule[0.8pt]
    
    \vspace{-15pt}
    \end{tabular*}

\end{wraptable}

Table~\ref{tab:pair_reconstruction} provides a diagnostic reconstruction
test. The singleton-additive approximation obtains an MAE of 0.018,
whereas including the exact measured pair interaction reduces the error
to 0.008, a 54.400\% reduction. Thus, second-order interactions account
for a substantial portion of the information that singleton scores fail
to capture.

Importantly, this experiment is diagnostic rather than part of the
deployed policy. DRSR never observes exact pair interactions online.
Instead, it learns set-level deletion risk from online-visible
candidate--state relations together with deleted, retained, and
pair-pooled representations.

Taken together, the poor singleton aggregation accuracy, the increasing
interaction magnitude with deletion-set cardinality, and the large
reconstruction improvement from pair information provide complementary
evidence for modeling deletion risk directly at the set level.

\FloatBarrier

\subsection{Full260 End-to-End Results}
\label{sec:full260}
Table~\ref{tab:full260} reports the principal Full260 result together with \emph{Structural Pruning}, our simple rule-based reference. Structural Pruning reaches a reported overall reward of 0.701 while reducing the token mean from 2.685M to 2.292M per task. DRSR improves substantially beyond this reference, reaching 0.802 reward at 2.126M tokens per task. The gain is especially pronounced in Security, where Structural Pruning reaches 0.640 and DRSR reaches 0.756.

\begin{table}[!htbp]
\centering
\caption{Full260 comparison by domain. $R$ denotes reward and $T$ denotes mean total tokens per task in millions. Overall reports the corresponding Full260 aggregate for each method.}
\label{tab:full260}
\tableformat
\fontsize{8.9}{10.7}\selectfont
\setlength{\tabcolsep}{1.6pt}
\renewcommand{\arraystretch}{1.08}
\begin{tabularx}{\linewidth}{@{}>{\raggedright\arraybackslash}p{0.255\linewidth}|YY|YY|YY|YY|YY@{}}
\toprule[0.9pt]
\multirow{2}{*}{\textbf{Method}} & \multicolumn{2}{c|}{\textbf{Code}} & \multicolumn{2}{c|}{\textbf{Office}} & \multicolumn{2}{c|}{\textbf{Security}} & \multicolumn{2}{c|}{\textbf{Web}} & \multicolumn{2}{c}{\textbf{Overall}} \\
\cmidrule(lr){2-3}\cmidrule(lr){4-5}\cmidrule(lr){6-7}\cmidrule(lr){8-9}\cmidrule(l){10-11}
& $R\uparrow$ & $T\downarrow$ & $R\uparrow$ & $T\downarrow$ & $R\uparrow$ & $T\downarrow$ & $R\uparrow$ & $T\downarrow$ & $R\uparrow$ & $T\downarrow$ \\
\midrule
\mbox{Baseline} & 0.770 & 1.370 & 0.818 & 1.198 & 0.478 & 5.977 & 0.721 & 2.429 & 0.699 & 2.685 \\
\mbox{Structural Pruning} & 0.703 & 1.380 & 0.814 & 0.893 & 0.640 & 4.961 & 0.659 & 2.046 & 0.701 & 2.292 \\
\textbf{DRSR} & \best{0.858} & 1.235 & \best{0.824} & 0.997 & \best{0.756} & 4.569 & \best{0.760} & 1.858 & \best{0.802} & 2.126 \\
\bottomrule[0.9pt]
\end{tabularx}
\end{table}

The domain breakdown shows that the gain is not produced by a single uniform compression effect. Security has the largest baseline context and the largest reward improvement, while Office changes only modestly. DRSR nevertheless reduces processed tokens in all four domains, consistent with selective history removal rather than a fixed context budget.

\subsection{Context-Management Comparison on Eval40}
\label{sec:comparison}
Table~\ref{tab:comparison} compares DRSR with the uncompressed agent, simple context controls, and published context-management methods on the same fixed Eval40 task manifest. Each domain contains ten tasks. We report reward on its native 0--1 scale and mean input-plus-output tokens per task in millions; auxiliary summarization-model tokens are included where applicable.

On Eval40, DRSR achieves 0.794 overall reward at 1.211M tokens per task, compared with 0.675 reward and 1.887M tokens for the uncompressed agent. This is a reward gain of 0.119 while reducing total tokens by 35.850\%. CoMem and PACE use fewer tokens but obtain substantially lower reward, while the strongest non-DRSR reward (Periodic Summary, $n=5$) remains below DRSR while using nearly twice as many tokens.

\FloatBarrier
\subsection{Component Ablation}
\label{sec:ablation}
We ablate the three design choices that distinguish DRSR from independent pruning: retained-context features, pair pooling inside the deleted set, and the abstention gate. Table~\ref{tab:ablation} reports reward and total token usage in each Eval40 domain, together with the equal-domain overall mean. Direct-removal statistics are reported in the discussion below and detailed in Appendix~\ref{app:domain_results}.

\begin{table}[!htbp]
\centering

\caption{
Eval40 context-management comparison.
$R$ is reward and $T$ is mean model tokens per task in millions.
}
\label{tab:comparison}

\tableformat
\fontsize{7.75}{9.25}\selectfont
\setlength{\tabcolsep}{0.55pt}
\renewcommand{\arraystretch}{1.03}

\begin{tabularx}{\linewidth}{
@{}
>{\raggedright\arraybackslash}p{0.315\linewidth}
|YY|YY|YY|YY|YY
@{}
}

\toprule[0.9pt]

\multirow{2}{*}{\textbf{Method}}
& \multicolumn{2}{c|}{\textbf{Code}}
& \multicolumn{2}{c|}{\textbf{Office}}
& \multicolumn{2}{c|}{\textbf{Security}}
& \multicolumn{2}{c|}{\textbf{Web}}
& \multicolumn{2}{c}{\textbf{Overall}}
\\

\cmidrule(lr){2-3}
\cmidrule(lr){4-5}
\cmidrule(lr){6-7}
\cmidrule(lr){8-9}
\cmidrule(l){10-11}

& $R\uparrow$ & $T\downarrow$
& $R\uparrow$ & $T\downarrow$
& $R\uparrow$ & $T\downarrow$
& $R\uparrow$ & $T\downarrow$
& $R\uparrow$ & $T\downarrow$
\\

\midrule

\mbox{Baseline}
& 0.729 & 1.438
& \second{0.816} & 1.040
& 0.446 & 3.973
& 0.710 & 1.098
& 0.675 & 1.887
\\

\doublerule

\mbox{Sliding Window ($K=5$)}
& 0.668 & 3.123
& 0.674 & 3.982
& 0.323 & 2.126
& 0.590 & 2.577
& 0.564 & 2.952
\\

\mbox{Sliding Window ($K=10$)}
& 0.744 & 1.756
& 0.731 & 1.816
& 0.401 & 0.819
& 0.610 & 3.547
& 0.621 & 1.985
\\

\mbox{Sliding Window ($K=20$)}
& 0.793 & 1.419
& \best{0.851} & 2.231
& 0.403 & 1.132
& 0.700 & 3.300
& 0.687 & 2.020
\\

\mbox{Periodic Summary ($n=3$)}
& 0.654 & 1.913
& 0.677 & 2.135
& 0.263 & 0.990
& \second{0.827} & 3.819
& 0.605 & 2.214
\\

\mbox{Periodic Summary ($n=5$)}
& \second{0.836} & 1.758
& 0.757 & 2.552
& 0.400 & 1.292
& 0.813 & 3.883
& \second{0.702} & 2.371
\\

\midrule

\mbox{PACE~\citep{wei2026pace}}
& 0.704 & 0.941
& 0.558 & 1.013
& 0.435 & 0.573
& 0.640 & 1.885
& 0.584 & \second{1.103}
\\

\mbox{LLMLingua-2~\citep{pan2024llmlingua}}
& 0.646 & 1.630
& 0.733 & 3.412
& 0.401 & 3.487
& 0.700 & 1.090
& 0.620 & 2.405
\\

\mbox{SelfCompact~\citep{li2026self}}
& 0.768 & 1.571
& 0.796 & 0.850
& 0.362 & 3.136
& 0.710 & 0.712
& 0.659 & 1.567
\\

\mbox{ACON-Core~\citep{kang2025acon}}
& 0.718 & 1.476
& 0.686 & 2.154
& \second{0.480} & 1.511
& 0.650 & 0.473
& 0.633 & 1.403
\\

\mbox{Self-GC~\citep{hao2026self}}
& 0.635 & 1.167
& 0.717 & 1.178
& 0.437 & 3.438
& 0.710 & 0.837
& 0.625 & 1.655
\\

\mbox{LRE~\citep{jahan2026learning}}
& 0.626 & 2.258
& 0.634 & 3.470
& 0.271 & 2.850
& 0.680 & 0.586
& 0.553 & 2.291
\\

\mbox{CoMem~\citep{zhang2026comem}}
& 0.564 & 0.550
& 0.599 & 0.745
& 0.025 & 1.004
& 0.580 & 0.764
& 0.442 & \best{0.766}
\\

\mbox{SAM~\citep{hu2026sam}}
& 0.676 & 1.412
& 0.816 & 1.094
& 0.472 & 1.910
& 0.670 & 0.969
& 0.659 & 1.346
\\

\mbox{SWE-Pruner~\citep{wang2026swe}}
& 0.635 & 2.172
& 0.726 & 1.431
& 0.409 & 1.822
& 0.650 & 0.582
& 0.605 & 1.502
\\

\mbox{Sculptor~\citep{li2026sculptor}}
& 0.512 & 1.029
& 0.816 & 0.985
& 0.353 & 2.453
& 0.690 & 0.708
& 0.593 & 1.294
\\

\mbox{ACM~\citep{li2026acm}}
& 0.387 & 2.540
& 0.559 & 1.526
& 0.062 & 5.196
& 0.340 & 0.193
& 0.337 & 2.364
\\

\doublerule

\textbf{DRSR}
& \best{0.856} & 1.216
& 0.788 & 0.724
& \best{0.700} & 1.616
& \best{0.833} & 1.288
& \best{0.794} & 1.211
\\

\bottomrule[0.9pt]

\end{tabularx}
\end{table}

\begin{table}[!htbp]
\centering
\caption{
Eval40 component ablation.
$R$ denotes reward and $T$ denotes mean total model tokens per task in
millions. Overall is the equal-domain mean. Direct is the mean number of tokens
directly removed per task in thousands across the fixed evaluation tasks.
}
\label{tab:ablation}

\tableformat
\fontsize{8.35}{9.95}\selectfont
\setlength{\tabcolsep}{0.85pt}
\renewcommand{\arraystretch}{1.06}

\begin{tabularx}{\linewidth}{
@{}
>{\raggedright\arraybackslash}p{0.185\linewidth}
|YY|YY|YY|YY|YYY
@{}
}
\toprule[0.9pt]

\multirow{2}{*}{\textbf{Variant}}
& \multicolumn{2}{c|}{\textbf{Code}}
& \multicolumn{2}{c|}{\textbf{Office}}
& \multicolumn{2}{c|}{\textbf{Security}}
& \multicolumn{2}{c|}{\textbf{Web}}
& \multicolumn{3}{c}{\textbf{Overall}}
\\

\cmidrule(lr){2-3}
\cmidrule(lr){4-5}
\cmidrule(lr){6-7}
\cmidrule(lr){8-9}
\cmidrule(l){10-12}

& $R\uparrow$ & $T\downarrow$
& $R\uparrow$ & $T\downarrow$
& $R\uparrow$ & $T\downarrow$
& $R\uparrow$ & $T\downarrow$
& $R\uparrow$ & $T\downarrow$
& \textbf{Direct}
\\

\midrule

\textbf{DRSR}
& \best{0.856} & 1.216
& \second{0.788} & 0.724
& \second{0.700} & 1.616
& \best{0.833} & 1.288
& \best{0.794} & 1.211
& 25.155
\\

\doublerule

w/o retained
& 0.722 & 0.716
& 0.779 & 0.266
& 0.698 & 1.254
& \second{0.800} & 1.731
& 0.750 & 0.992
& 12.633
\\

w/o pair pooling
& 0.756 & 0.567
& 0.678 & 0.606
& \best{0.796} & 1.094
& 0.796 & 2.535
& \second{0.757} & 1.201
& 32.205
\\

w/o abstention
& \second{0.833} & 1.205
& \best{0.801} & 0.447
& 0.573 & 0.963
& \second{0.800} & 1.553
& 0.752 & 1.042
& 30.093
\\

\bottomrule[0.9pt]
\end{tabularx}
\end{table}

The ablation effects are domain dependent, but every removal lowers the overall reward. Removing abstention increases direct deletion from 25.155K to 30.093K tokens per task while lowering overall reward from 0.794 to 0.752. Removing retained-context features makes the policy more conservative in direct deletion (12.633K) yet still lowers overall reward to 0.750, indicating that knowing what survives the edit matters for identifying redundancy. Removing pair pooling yields the largest direct deletion, 32.205K tokens per task, while overall reward falls to 0.757. Together with the measured interaction residuals, this pattern is consistent with multi-Block deletions appearing safer when within-set interaction is not represented.

\section{Conclusion}
\label{sec:conclusion}
We formulate agent-history pruning as selection over the joint deletion risk of
structured Blocks. DRSR learns this risk from offline counterfactual supervision
and uses a compact scorer to gate online pruning with abstention. Our analyses
show that historical relevance is decision-conditioned and joint deletion effects
cannot be recovered reliably from independently scored Blocks. On WorkBuddyBench
Full260, DRSR improves reward from 0.699 to 0.802 while reducing tokens by
20.820\%; on Eval40, it reaches 0.794 reward with 35.850\% fewer tokens. These
results support state-dependent, set-level compression over independent
importance ranking of historical units. More broadly, the results indicate that
safe context reduction requires modeling both the current decision state and the
joint structure of candidate deletions. This perspective provides a practical
basis for reducing long-horizon agent context without relying on fixed
state-independent importance scores.

\FloatBarrier
\section*{Reproducibility Statement}
The deletion action space, counterfactual harm target, and label threshold are defined in Section~\ref{sec:methodology}; candidate construction, relational features, set representation, training objective, and online selection are specified in Section~\ref{sec:drsr}. Appendix~\ref{app:algorithm} gives the complete decision procedure, Appendix~\ref{app:features} lists the representation and scorer configuration, Appendix~\ref{app:optimization} defines the training and evaluation metrics, Appendix~\ref{app:transition_heuristic} documents the simple structural baseline, Appendix~\ref{app:domain_results} reports detailed token summaries, and Appendix~\ref{app:deployment_details} specifies the information boundary and validation-frozen deployment rule.

\FloatBarrier
\section*{AI Use Statement}
Generative AI was used to aid or polish writing and for retrieval and discovery. This included improving language, readability, clarity, organization, and \LaTeX{} formatting, as well as assisting with literature discovery. All AI-assisted text and retrieved references were carefully reviewed, verified, and approved by the authors, who take full responsibility for the entire submission.

\FloatBarrier
\IfFileExists{iclr2027_conference.bst}{%
  \bibliographystyle{unsrtnat}
}{%
  \bibliographystyle{unsrtnat}
}
\begingroup
\fontsize{9}{10.8}\selectfont
\setlength{\bibsep}{3pt plus 1pt}
\clearpage
\setlength{\bibsep}{.5ex plus .8ex}
\bibliography{iclr2027_conference}
\endgroup

\clearpage
\clearpage
\appendix

\setlength{\textfloatsep}{9pt plus 2pt minus 2pt}
\setlength{\floatsep}{8pt plus 2pt minus 2pt}
\setlength{\intextsep}{8pt plus 2pt minus 2pt}

\section{Decision Procedure}
\label{app:algorithm}

\begin{algorithm}[!htbp]
\caption{One DRSR decision}
\label{alg:drsr}
\begin{algorithmic}[1]
\Require Valid current request $C_t$; frozen representation model; scorer ensemble; $K=3$, $q_{\max}=0.20$, $\tau=0.0740822119017442$
\Ensure Protocol-valid pruned request or unchanged $C_t$
\State Protect the system/task prefix and Recent2; build an aligned representation view within the context cap.
\State Construct outcome-blind protocol-valid candidates $E_t$.
\If{fewer than three candidates, alignment failure, or invalid protocol}
    \State \Return $C_t$
\EndIf
\State Extract structural, attention, hidden-state, and projected semantic features from the intact representation view.
\State Enumerate all nonempty $A\subseteq E_t$ with $|A|\le K$ and batch-score them to obtain $\bar H_t(A)$ and $U_t(A)$.
\State Map each subset to the full request, measure $w_t(A)$, and form the feasible set using protocol, recency, removal-budget, and risk constraints.
\If{no feasible set remains}
    \State \Return $C_t$
\EndIf
\State Select $A_t^\star$ by maximum direct saving, breaking ties by lower unsafe score and then lower predicted harm.
\State Jointly delete $A_t^\star$, reserialize the request, and validate the final tool protocol.
\State \Return the valid rewritten request; otherwise return $C_t$.
\end{algorithmic}
\end{algorithm}

\section{Supervision and Task Sets}
\label{app:setup}

\begin{table}[!htbp]
\centering
\caption{Offline supervision and benchmark task counts.}
\label{tab:data}
\tableformat
\fontsize{9.5}{11.2}\selectfont
\setlength{\tabcolsep}{7pt}
\begin{tabular*}{0.92\linewidth}{@{\extracolsep{\fill}}lccccc@{}}
\toprule[0.9pt]
\multicolumn{6}{c}{\textit{External-trajectory supervision}} \\
\midrule
\textbf{Source trajectories} & \textbf{Decision states} & \textbf{Set-level labels} & \multicolumn{3}{c}{\textbf{Train / validation / test}} \\
566 & 1,698 & $\approx24{,}122$ & \multicolumn{3}{c}{80:10:10 by trajectory} \\
\doublerule
\multicolumn{6}{c}{\textit{Task evaluation}} \\
\midrule
\textbf{Set} & \textbf{Code} & \textbf{Office} & \textbf{Security} & \textbf{Web} & \textbf{Total} \\
Full260 & 80 & 50 & 60 & 70 & 260 \\
Eval40 & 10 & 10 & 10 & 10 & 40 \\
\bottomrule[0.9pt]
\end{tabular*}
\tablenote{The supervision set contains approximately 24,122 set-level deletion labels and is split 80:10:10 by source trajectory.}
\end{table}

All states and deletion sets generated from one source trajectory remain in the same split. This grouping prevents closely related deletion rows from crossing train, validation, and test partitions. Identical task counts do not by themselves establish a paired benchmark comparison. Paired evaluation additionally requires the same task manifest, acting-model configuration, and token-accounting rules.

The numerical value $0.020$ appears in two different roles. The meaningful-gap threshold selects which pairs are counted in pair-ordering evaluation, while $\delta_H$ in Eq.~\eqref{eq:label} defines the offline unsafe event. The two tests use the same numerical margin but serve different purposes.

\section{Representation and Scorer Details}
\label{app:features}

\subsection{Candidate Construction and Structural Controls}
\label{app:candidate_features}

Candidate construction is frozen before any deletion outcome is observed.
The two most recent complete Steps are protected. Earlier protocol-valid
atoms are grouped into at most four chronological bins according to the
position of their parent Step, and the longest atom in each bin is selected.
This produces three or four candidates for an eligible request. If fewer
than three candidates remain, alignment fails, or a rewrite is invalid,
the request is left unchanged.

For each candidate Block $B_i$, structural controls describe properties
that are directly visible before the next action. They include parent-Step
age and relative position, Block type and component index, character count,
word count, removable token count, lexical overlap with the task and recent
history, tool-related overlap, redundancy with other candidates, and simple
content indicators such as error, success, path, numeric, and nonzero-exit
density. These controls provide context for the learned scorer. They are
not used as a standalone importance rule.

Representation extraction uses the frozen Qwen3.5-9B backbone. If the full
request exceeds 55,296 tokens, DRSR takes the longest exactly aligned suffix
that still contains Recent2. Candidate spans extracted from this view are
mapped back to the complete serialized request before any token-saving or
removal constraint is evaluated.

\subsection{Attention Relations}
\label{app:attention_features}

Let $t$ denote the final pre-action position in the intact causal prefix.
For attention layer $\ell$, head $h$, and historical token position $k$,
the standard attention weight is
\begin{equation}
\alpha_{t,k}^{\ell,h}
=
\operatorname{softmax}_{k}
\left(
\frac{
q_t^{\ell,h}
(k_k^{\ell,h})^\top
}{
\sqrt{d_h}
}
\right).
\label{eq:attention_weight}
\end{equation}
For the token span $\Omega_i$ of Block $B_i$, DRSR uses the attention mass
and density
\begin{equation}
a_{i,t}^{\ell,h}
=
\sum_{k\in\Omega_i}
\alpha_{t,k}^{\ell,h},
\qquad
d_{i,t}^{\ell,h}
=
\frac{a_{i,t}^{\ell,h}}
{|\Omega_i|}.
\label{eq:attention_app}
\end{equation}
Mass measures the total attention assigned to the Block, while density
normalizes that quantity by Block length. The implementation records
head-level mass and density together with summary statistics across heads,
including mean and maximum mass, mean and maximum density, and the mean of
the largest token-level attention values within the Block.

Formal attention features are extracted only from the standard
full-attention layers
\[
\ell\in\{4,8,12,16,20,24,28,32\}.
\]
Other backbone layers use a different attention implementation and are not
treated as measurements of the same attention quantity. Any interpolated
points used only for visualization are excluded from the formal features
and from the reported attention analysis.

Attention is not interpreted as a deletion score. A Block can receive low
attention and still carry a constraint whose removal changes the next
decision. Conversely, a highly attended Block can be redundant when similar
evidence remains elsewhere. Attention therefore enters the candidate
representation only as one online-visible relation.

\subsection{Hidden-State Relations}
\label{app:hidden_features}

The hidden-state features are designed to measure how a historical Block
relates to the current decision state. They do not ask whether the Block has
high value in isolation.

Let $h_k^\ell\in\mathbb R^{d_\ell}$ be the hidden state of token $k$ at
layer $\ell$. For Block $B_i$ with token span $\Omega_i$, define its mean
hidden representation as
\begin{equation}
\bar h_i^\ell
=
\frac{1}{|\Omega_i|}
\sum_{k\in\Omega_i}
h_k^\ell,
\label{eq:block_hidden_mean}
\end{equation}
and let $h_t^\ell$ denote the hidden state at the final pre-action position.
DRSR compares the two representations through
\begin{equation}
\cos(\bar h_i^\ell,h_t^\ell),
\qquad
\frac{
\|\bar h_i^\ell-h_t^\ell\|_2
}{
\sqrt{d_\ell}
},
\qquad
\log
\frac{
\|\bar h_i^\ell\|_2
}{
\|h_t^\ell\|_2
}.
\label{eq:hidden_features}
\end{equation}
The cosine term measures directional similarity, the normalized Euclidean
distance measures separation in representation space, and the log norm
ratio records differences in representation magnitude without allowing
hidden dimension to dominate the scale.

Mean-span features can hide information concentrated at the boundaries of
a Block. The implementation therefore also includes cosine relations
between $h_t^\ell$ and the first and last token states of the Block:
\begin{equation}
\cos(h_{i,\mathrm{first}}^\ell,h_t^\ell),
\qquad
\cos(h_{i,\mathrm{last}}^\ell,h_t^\ell).
\label{eq:hidden_boundary}
\end{equation}
Together, the mean and boundary relations provide several views of the
same state-conditioned question: whether the representation of historical
content remains aligned with the representation immediately before the
next action.

These quantities are descriptive features, not causal importance measures.
Their usefulness is learned from offline deletion labels. In particular,
a large cosine similarity does not force retention, and a large distance
does not imply safe deletion. Numerical implementations stabilize zero
norms consistently before cosine and norm-ratio computation.

\subsection{Fixed Projected Semantic Relations}
\label{app:semantic_features}

Scalar similarities compress a high-dimensional relation into a small
number of values. DRSR therefore retains additional directional
information through fixed random projections. At semantic layers
\[
\ell\in\{8,16,24,32\},
\]
let
$R_\ell\in\mathbb R^{d_\ell\times64}$ be a Gaussian projection whose
entries are drawn independently with variance $1/64$. The projection seed
is fixed to $73000+\ell$. The projected Block and current-state pair is
\begin{equation}
s_{i,t}^{\ell}
=
\left[
\frac{\bar h_i^\ell}{\|\bar h_i^\ell\|_2}R_\ell,\,
\frac{h_t^\ell}{\|h_t^\ell\|_2}R_\ell
\right].
\label{eq:projection}
\end{equation}
The projection matrices are fixed before training and are independent of
all deletion labels. They therefore provide a compact view of semantic
directions without fitting the representation backbone or the projection
to downstream outcomes.

\subsection{Candidate Feature Assembly}
\label{app:candidate_assembly}

The complete candidate vector concatenates the structural, attention,
hidden-state, and projected semantic components:
\begin{equation}
x_{i,t}
=
\left[
x_{i,t}^{\mathrm{struct}},
x_{i,t}^{\mathrm{attn}},
x_{i,t}^{\mathrm{hidden}},
x_{i,t}^{\mathrm{proj}}
\right]
\in\mathbb R^{1154}.
\label{eq:candidate_feature}
\end{equation}
Every component is computed from the intact causal prefix. No candidate
deletion is executed to construct $x_{i,t}$, and no realized future output,
measured harm, or task reward is included. The feature order is fixed
across training and deployment.

The candidate features retain temporal information through structural
position variables even though the later set encoder is permutation
invariant within the deleted and retained groups. The representation view
can be shorter than the complete request because of the 55,296-token cap,
but token saving, Recent2 protection, protocol validation, and removal
limits are always evaluated after mapping candidate spans back to the full
request.

\subsection{Exact Set Representation}
\label{app:exact_setrep}

A two-layer Linear/GELU encoder maps the 1,154-dimensional candidate vector
to a 64-dimensional embedding
\begin{equation}
e_i
=
\phi_\theta(x_{i,t})
\in\mathbb R^{64}.
\label{eq:unit_encoder}
\end{equation}
For deleted candidates $D=A$ and retained candidates $R=E_t\setminus A$,
let $\mu_G$ and $m_G$ denote the mean and elementwise maximum of the
embeddings in a nonempty group $G$. Empty groups use zero-padded summaries.

The deleted-set pair statistic is
\begin{equation}
p_D
=
\frac{
\left(\sum_{i\in D}e_i\right)^{\odot2}
-
\sum_{i\in D}e_i^{\odot2}
}{
2\max\left\{\binom{|D|}{2},1\right\}
}.
\label{eq:pairpool}
\end{equation}
For $|D|\ge2$, this equals the average elementwise product over unordered
pairs of deleted candidates. For a singleton, the numerator is zero and
$p_D$ is the zero vector. Appendix~\ref{app:properties} gives the algebraic
identity behind this pooling operation.

The model also compares the deleted and retained groups through
\begin{equation}
c_{D,R}
=
\mu_D\odot\mu_R,
\qquad
g_{D,R}
=
|\mu_D-\mu_R|.
\label{eq:deleted_retained_geometry}
\end{equation}
These terms expose whether the proposed deletion and the surviving
candidate information occupy similar or different regions of the learned
candidate space. They do not impose a hand-written redundancy rule. Their
effect is learned from the set-level supervision.

The budget vector is
\begin{equation}
b_t(A)
=
\left[
\frac{|A|}{|E_t|},\,
\frac{\log(1+w_t(A))}{10},\,
\frac{w_t(A)}
{\sum_{B_i\in E_t}w_t(\{B_i\})}
\right].
\label{eq:budget}
\end{equation}
The exact set representation is
\begin{equation}
r_t^{\mathrm{exact}}(A)
=
[
\mu_D,\mu_R,m_D,m_R,p_D,
\mu_D\odot\mu_R,
|\mu_D-\mu_R|,
b_t(A)
]
\in\mathbb R^{451}.
\label{eq:setrep_exact}
\end{equation}
Each of the seven pooled embedding terms contributes 64 coordinates and
$b_t(A)$ contributes three, giving $7\times64+3=451$ features.

The pair statistic provides information about interactions inside the
deleted set, while the deleted and retained geometry provides information
about what survives after the edit. The final scorer is not restricted to
an additive sum of singleton or pair effects.

\subsection{Scorer and Training Objective}
\label{app:scorer_training}

A shared 64-dimensional trunk maps
$r_t^{\mathrm{exact}}(A)$ to $h_A$. Cardinality-indexed heads produce
\begin{equation}
\widehat H_t(A)
=
g^{\mathrm{harm}}_{|A|}(h_A),
\qquad
\widehat p_t(A)
=
\sigma
\left(
g^{\mathrm{risk}}_{|A|}(h_A)
\right).
\label{eq:scorer_heads}
\end{equation}
Training includes deletion cardinalities one through four, although online
selection permits at most three deleted candidates. Separate heads allow
the output mapping to vary with set size, which is useful because the scale
and frequency of interaction can change with cardinality.

The two outputs have different roles. The harm head learns a continuous
quantity and supports ranking among candidate sets. The unsafe head learns
the event in Eq.~\eqref{eq:label} that is used by the online gate. The
unsafe output is a learned decision score and is not interpreted as a
calibrated probability of downstream task failure.

The composite training objective is
\begin{equation}
\mathcal L_{\mathrm{train}}
=
\mathcal L_{\mathrm{rank}}
+
0.30\mathcal L_{\mathrm{SmoothL1}}
+
0.40\mathcal L_{\mathrm{BCE}}
+
0.10\mathcal L_{\mathrm{Brier}}.
\label{eq:loss}
\end{equation}
The regression term fits measured joint harm. Weighted BCE uses
\begin{equation}
w_+
=
\min\left(\frac{N_-}{N_+},4\right),
\label{eq:bce_weight}
\end{equation}
and the Brier term penalizes squared error on the unsafe target. The ranking
term is defined in Appendix~\ref{app:optimization}. Checkpoint selection
lexicographically prioritizes unsafe AUROC, meaningful-gap pair accuracy,
and negative Brier score. The test split is used for neither fitting nor
threshold selection.

\Needspace{360pt}
\subsection{Model and Training Configuration}
\label{app:model_configuration}

\begin{table}[H]
\centering
\caption{Model and training configuration. The language models are frozen; only the set-risk scorer is optimized.}
\label{tab:configuration}
\tableformat
\setlength{\tabcolsep}{10pt}
\begin{tabular*}{0.86\linewidth}{@{\extracolsep{\fill}}ll@{}}
\toprule[0.9pt]
\textbf{Configuration item} & \textbf{Value} \\
\midrule
\multicolumn{2}{l}{\textit{Representation and scorer}} \\[1pt]
Representation backbone & Frozen Qwen3.5-9B \\
Acting model & DeepSeek-V4-Flash \\
Candidate features / embedding & 1,154 / 64 dimensions \\
Set representation / shared hidden state & 451 / 64 dimensions \\
Attention layers & 4, 8, 12, 16, 20, 24, 28, 32 \\
Semantic layers / projection size & 8, 16, 24, 32 / 64 \\
Projection seed at layer $\ell$ & $73000+\ell$ \\
Context / target token cap & 55,296 / 3,072 \\
\doublerule
\multicolumn{2}{l}{\textit{Scorer optimization}} \\[1pt]
Optimizer / learning rate & AdamW / $2\times10^{-3}$ \\
Weight decay / gradient clipping & $10^{-4}$ / 5.0 \\
Maximum epochs / evaluation interval & 300 / every 5 epochs \\
Early-stopping patience & 12 evaluations \\
Ensemble seeds & 11, 29, 47 \\
\doublerule
\multicolumn{2}{l}{\textit{Online constraints}} \\[1pt]
Maximum candidates / deletion cardinality & 4 / 3 \\
Protected complete Steps & Most recent 2 \\
Per-request removal cap & 20.000\% of full request \\
Harm-label threshold $\delta_H$ & 0.020 \\
Deployment threshold $\tau$ & 0.0740822119017442 \\
Uncertainty coefficient $\beta$ & 0 \\
No feasible deletion set & Retain the complete request \\
\bottomrule[0.9pt]
\end{tabular*}
\end{table}

\section{Algebraic Properties and Design Implications}
\label{app:properties}

This section collects identities used by the main formulation and clarifies
what they do and do not imply. These results are algebraic properties of the
defined quantities. They are not claims that the learned scorer recovers the
true harm exactly.

\subsection{Telescoping Decomposition of Joint Harm}
\label{app:telescoping_proof}

For an ordered deletion sequence
$A_k=\{B_{i_1},\ldots,B_{i_k}\}$ with $A_0=\varnothing$, recall
\[
H_t(B_{i_k}\mid A_{k-1})
=
H_t(A_k)-H_t(A_{k-1}).
\]

\paragraph{Proposition 1.}
If all intermediate rewrites are valid, then
\begin{equation}
H_t(A_m)
=
\sum_{k=1}^{m}
H_t(B_{i_k}\mid A_{k-1}).
\label{eq:telescoping_app}
\end{equation}

\paragraph{Proof.}
Substituting the definition of the conditional effect gives
\begin{align}
\sum_{k=1}^{m}
H_t(B_{i_k}\mid A_{k-1})
&=
\sum_{k=1}^{m}
\left[
H_t(A_k)-H_t(A_{k-1})
\right] \\
&=
H_t(A_m)-H_t(A_0).
\end{align}
Since $A_0=\varnothing$ and deleting the empty set leaves the context
unchanged, $H_t(A_0)=0$. This yields Eq.~\eqref{eq:telescoping_app}.
\hfill$\square$

The identity explains why singleton harm is insufficient in general.
Each conditional term is evaluated after a different subset has already
been removed. The final joint harm is independent of the chosen deletion
order, but the intermediate conditional effects need not be.

\subsection{Singleton Additivity and Interaction Residuals}
\label{app:interaction_properties}

For a set $A$, the interaction residual is
\[
I_t(A)
=
H_t(A)
-
\sum_{B_i\in A}H_{i,t}.
\]
Therefore
\begin{equation}
H_t(A)
=
\sum_{B_i\in A}H_{i,t}
+
I_t(A).
\label{eq:interaction_decomposition}
\end{equation}
Exact singleton additivity for a specific set $A$ holds if and only if
$I_t(A)=0$. This statement is set specific. Observing zero residual for one
set does not imply that larger or different sets are additive.

For two Blocks,
\begin{equation}
H_t(\{B_i,B_j\})
=
H_{i,t}
+
H_{j,t}
+
J_{ij,t},
\label{eq:pair_reconstruction_exact}
\end{equation}
where $J_{ij,t}$ is defined in Eq.~\eqref{eq:interaction}. Exact pair
interaction therefore reconstructs pair harm by definition. For sets with
three or more Blocks, pair terms alone need not recover all higher-order
effects. This is why the diagnostic pair reconstruction experiment does
not become the deployed prediction rule.

\subsection{Why Retained Context Changes Deletion Risk}
\label{app:retained_context}

For $B_i\notin A$,
\begin{equation}
H_t(B_i\mid A)
=
H_t(A\cup\{B_i\})
-
H_t(A).
\label{eq:conditional_effect_app}
\end{equation}
The effect of deleting $B_i$ is therefore indexed by the set already removed,
or equivalently by the context that remains. If another Block contains
redundant evidence, deleting $B_i$ can have small conditional effect while
that evidence remains and a much larger effect after the redundant source is
also removed.

This dependence motivates the deleted and retained partition in the set
encoder. A model that receives only representations of the deleted Blocks
cannot directly distinguish two proposals with the same deleted content but
different surviving evidence. DRSR therefore includes explicit retained-set
summaries and deleted-to-retained geometry.

\subsection{Identity Behind the Pair Pooling Statistic}
\label{app:pairpool_proof}

Let
\[
S_D=\sum_{i\in D}e_i.
\]
Elementwise squaring gives
\begin{equation}
S_D^{\odot2}
=
\sum_{i\in D}e_i^{\odot2}
+
2\sum_{\substack{i<j\\i,j\in D}}
e_i\odot e_j.
\label{eq:pair_expand}
\end{equation}
Rearranging,
\begin{equation}
\frac{
S_D^{\odot2}
-
\sum_{i\in D}e_i^{\odot2}
}{
2
}
=
\sum_{\substack{i<j\\i,j\in D}}
e_i\odot e_j.
\label{eq:pair_sum_identity}
\end{equation}
Dividing by $\binom{|D|}{2}$ when $|D|\ge2$ yields the average unordered
pair product used in Eq.~\eqref{eq:pairpool}. For $|D|=1$, the numerator is
zero, so the implementation defines $p_D$ as the zero vector.

This statistic is permutation invariant and can be computed without
explicitly listing every pair. It supplies information about pair structure
to the downstream network, but it does not impose a pairwise additive model
of harm.

\subsection{Oracle Constraint and Learned Online Gate}
\label{app:oracle_gate}

The ideal objective in Eq.~\eqref{eq:oracle_selection} uses the true event
\[
H_t(A)\le\delta_H.
\]
This event cannot be evaluated at online decision time because it depends on
the realized next output $y_t$. DRSR instead learns an unsafe score
$\widehat p_t(A)$ from offline labels and aggregates an ensemble score
$U_t(A)$ for deployment.

The online condition
\[
U_t(A)\le\tau
\]
should therefore be interpreted as a learned surrogate for feasibility, not
as an algebraic replacement for $H_t(A)\le\delta_H$. The two thresholds act
on different variables. The value $\delta_H$ defines the offline target,
while $\tau$ is selected on validation predictions under the policy
constraints described in Appendix~\ref{app:calibration}.

\subsection{Accumulation Without Interaction}
\label{app:accumulation}

Set-level reasoning is required even when the interaction residual is zero.
Suppose $I_t(A)=0$. Then
\[
H_t(A)
=
\sum_{B_i\in A}H_{i,t}.
\]
It is possible for every singleton to satisfy
$H_{i,t}\le\delta_H$ while the sum exceeds $\delta_H$. Thus independent
thresholding of singleton harms does not guarantee that their union is safe.
The cardinality cap and the direct set-level gate address this accumulation
failure in addition to nonadditive interaction.

\section{Structural Pruning Baseline}
\label{app:transition_heuristic}

\paragraph{Structural Pruning.}
We retain the earliest structural pruning implementation as a deliberately simple reference point. The heuristic protects the two most recent complete Steps, preserves tool calls and matched results, preserves state-bearing content, and removes only short transition narration for which later structure provides a witness. It has no learned risk estimator, no counterfactual supervision, and no retrieval or recall component.

On Full260, the supplied domain rewards are 0.703, 0.814, 0.640, and 0.659 for Code, Office, Security, and Web, respectively, with a reported overall reward of 0.701. Its mean total tokens per task are 1.380M, 0.893M, 4.961M, and 2.046M by domain, with 2.292M overall. At the aggregate level, this corresponds to a 14.640\% reduction from the canonical Full260 baseline. The independent table is omitted because the same reward and token values already appear in the main Full260 comparison.

\section{Additional Representation Results}
\label{app:extra_figures}

\begin{table}[!htbp]
\centering
\caption{History granularity and relational representations. Accuracy concerns the ordering of deletion effects, not task completion.}
\label{tab:representations}
\tableformat
\setlength{\tabcolsep}{9pt}
\begin{tabular*}{0.84\linewidth}{@{\extracolsep{\fill}}lr@{}}
\toprule[0.9pt]
\textbf{Measurement / representation} & \textbf{Value} \\
\midrule
\multicolumn{2}{l}{\textit{History granularity}} \\[1pt]
Median within-state Step-harm range & 0.312 \\
Median within-state Block-harm range & 0.028 \\
Unsafe parent Steps containing a safe Block & 71.400\% \\
\doublerule
\multicolumn{2}{l}{\textit{Controls: pair accuracy at harm margin 0.020}} \\[1pt]
Current state only & 50.000\% \\
Surface features & 64.750\% \\
\doublerule
\multicolumn{2}{l}{\textit{State relations: pair accuracy at harm margin 0.020}} \\[1pt]
Pre-action relation with matched state & 70.920\% \\
Pre-action relation with mismatched state & 62.020\% \\
Full-context relation & 66.390\% \\
Attention mass, layer 20 & 71.310\% \\
Combined relations, layers 12 and 16 & \best{72.130\%} \\
\bottomrule[0.9pt]
\end{tabular*}
\end{table}


\begin{figure}[!htbp]
\centering
\includegraphics[width=0.94\linewidth]{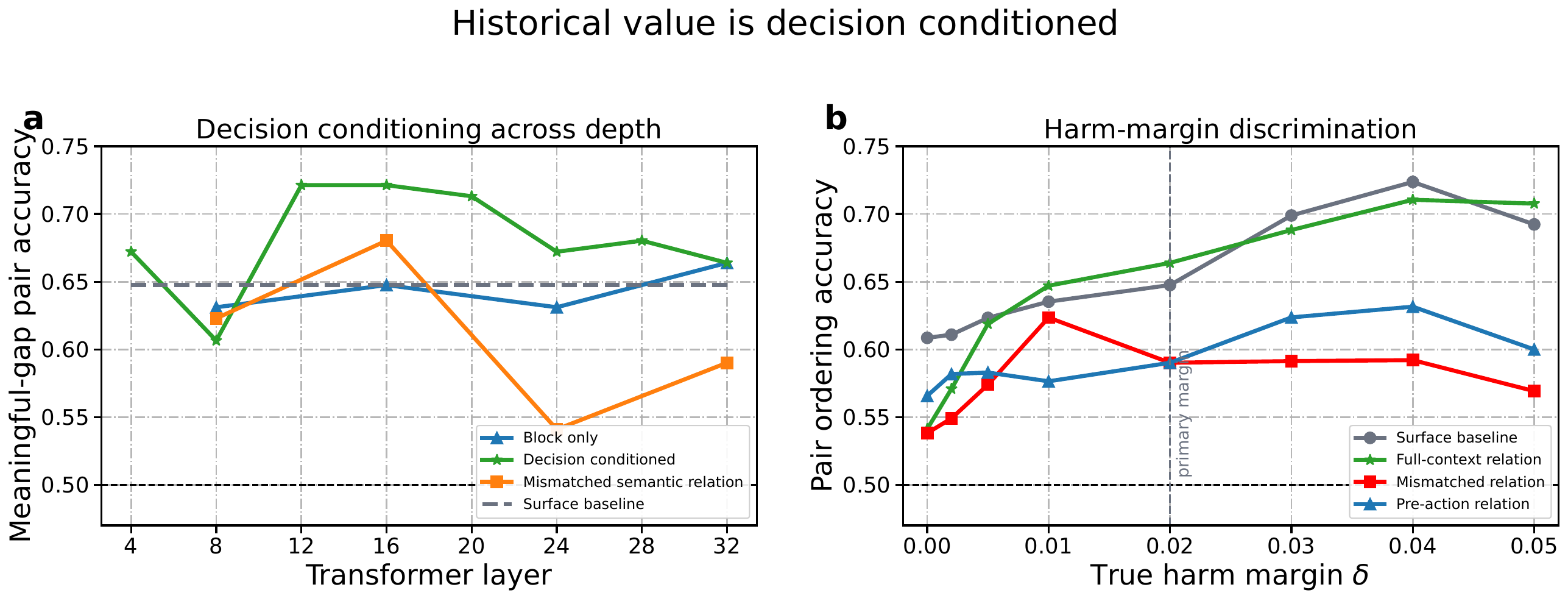}
\caption{
Decision-conditioned historical relevance.
(a) Meaningful-gap pair accuracy across representation depth for
decision-conditioned, Block-only, and mismatched-state relations,
with the surface-feature baseline for reference.
(b) Pair-ordering accuracy as the true harm margin varies; the dashed
line marks the primary evaluation margin.
}
\label{fig:conditioning_appendix}
\end{figure}

Figure~\ref{fig:conditioning_appendix} examines whether historical relevance
is better represented relative to the current decision state than from the
historical Block alone. At the primary harm margin, the matched pre-action
relation reaches 70.920\% pair accuracy, compared with 62.020\% when the same
candidate is paired with a mismatched state. The best combined relation reaches
72.130\%. The layerwise view further shows where this decision-conditioned
signal emerges across representation depth, while the margin sweep verifies
that the comparison is not specific to a single ordering threshold. Together,
these results support representing historical content through its relation to
the current state rather than assigning a fixed state-independent importance
score.


\begin{figure}[!htbp]
\centering
\includegraphics[width=0.94\linewidth]{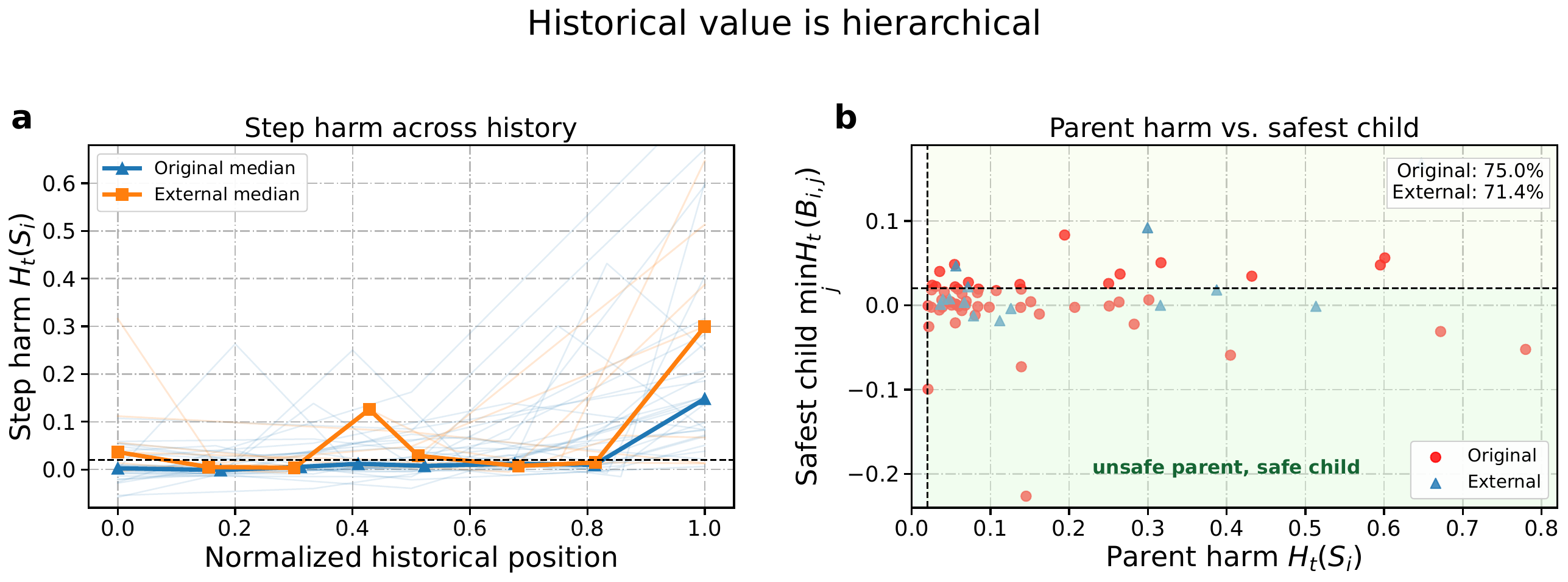}
\caption{History granularity analysis. (a) Step harm across normalized historical position. (b) Parent-Step harm against the lowest-harm constituent Block; the unsafe-parent/safe-child region shows why Block-level actions can preserve useful parts of an otherwise unsafe Step.}
\label{fig:hierarchy_appendix}
\end{figure}

Figure~\ref{fig:hierarchy_appendix} motivates the Block-level action space used
by DRSR. Step-level deletion aggregates heterogeneous assistant content and tool
exchanges into a single removal decision, producing a substantially wider
within-state harm range than Block-level deletion (0.312 versus 0.028).
Moreover, 71.400\% of unsafe parent Steps contain at least one constituent
Block that is individually safe to remove. The parent--child comparison in
panel (b) therefore shows why deleting an entire Step can discard useful
content unnecessarily: a coarse unit may be unsafe even when a finer-grained
subcomponent remains removable.


\begin{figure}[!htbp]
\centering
\includegraphics[width=\linewidth]{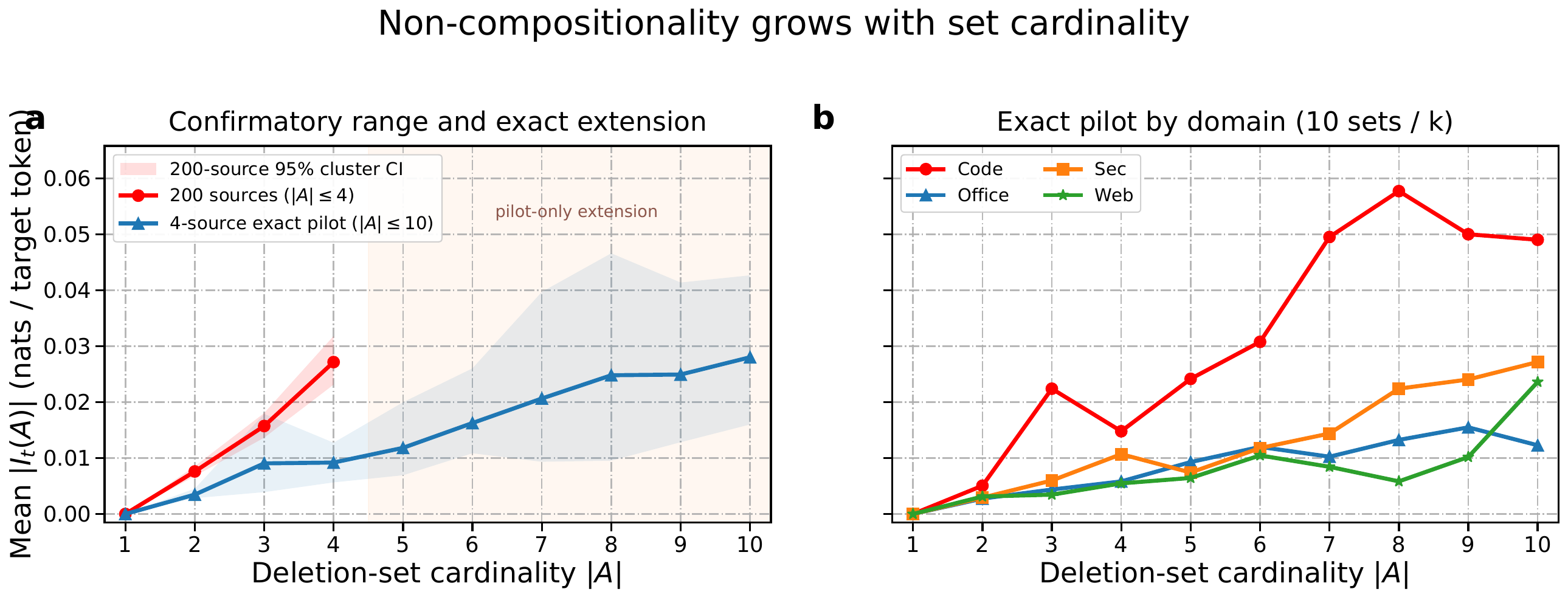}
\caption{Full non-compositionality analysis. The main paper uses only the 200-source confirmatory range $|A|\le4$; this appendix figure retains the exploratory exact extension to larger sets and the per-domain pilot.}
\label{fig:cardinality_full}
\end{figure}

Figure~\ref{fig:cardinality_full} extends the main-text cardinality analysis.
Within the confirmatory range, mean absolute non-compositionality increases
from 0 for singleton deletion to 0.008, 0.016, and 0.027 for deletion sets of
size two, three, and four. The discrepancy between independently measured
singleton effects and actual joint deletion therefore grows as more Blocks are
removed together. At cardinality four, its mean magnitude already exceeds the
$\delta_H=0.020$ harm threshold used to define unsafe supervision. The
additional larger-set and per-domain panels are retained as exploratory
diagnostics and are not used to define the deployed policy.

\subsection{Composition, Risk Scores, and Attention}


\begin{figure}[!htbp]
\centering
\includegraphics[width=\linewidth]{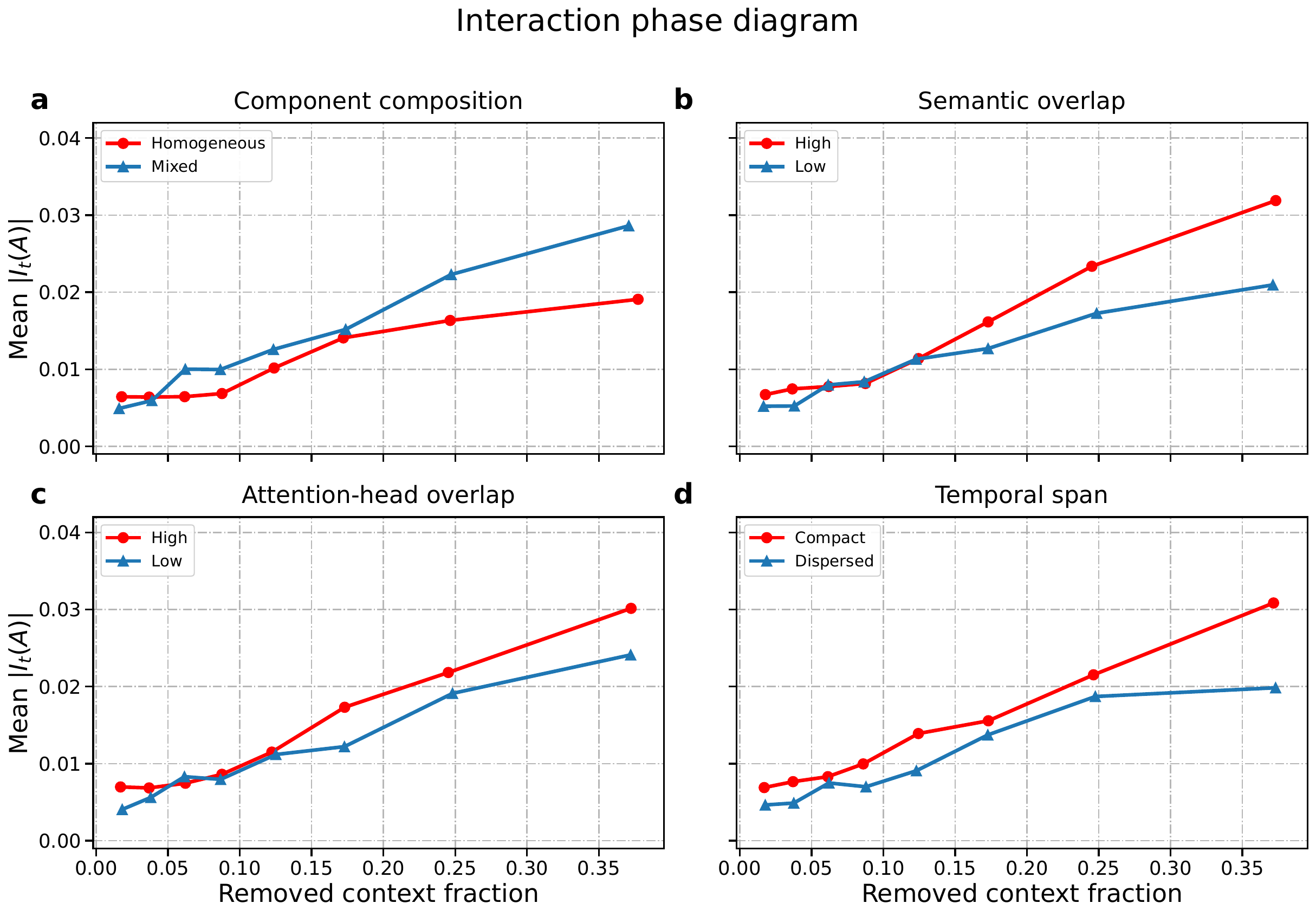}
\caption{Interaction summaries by component composition, semantic overlap, attention-head overlap, and temporal span.}
\label{fig:composition}
\end{figure}

Figure~\ref{fig:composition} decomposes the measured interaction residual along
several observable properties of a deletion set. The panels distinguish
component composition, semantic overlap, attention-head overlap, and temporal
separation, allowing the non-compositional effect to be inspected beyond set
cardinality alone. These analyses are diagnostic: they test whether joint harm
is associated with multiple forms of relation among historical Blocks, rather
than defining hand-written deletion rules. DRSR therefore learns from the
measured set-level labels instead of thresholding any individual overlap or
distance statistic.


\begin{figure}[!htbp]
\centering
\includegraphics[width=\linewidth]{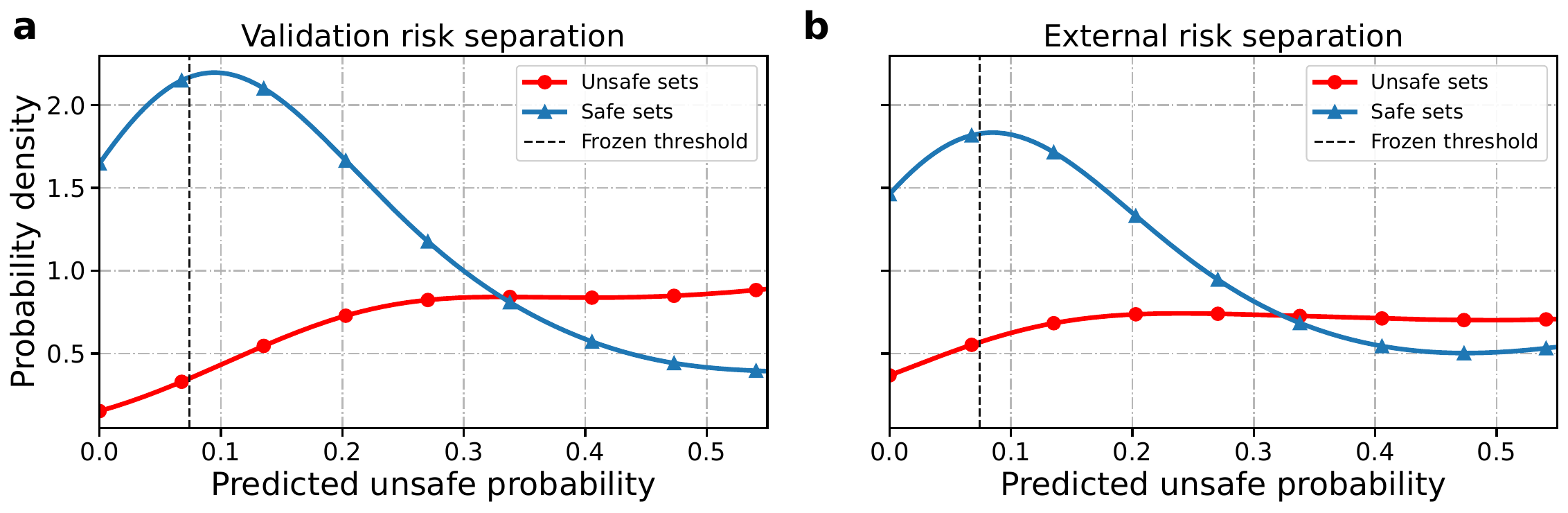}
\caption{Unsafe-score distributions for validation and external sets. The dashed line marks the deployment threshold; distributional separation is not a calibration guarantee.}
\label{fig:risk_distributions}
\end{figure}

Figure~\ref{fig:risk_distributions} shows the distribution of the learned
unsafe score on validation and external examples. The dashed line marks the
frozen deployment threshold $\tau$, making explicit how candidate sets are
partitioned by the online risk gate. The figure is intended as a distributional
diagnostic rather than a probability-calibration result: the unsafe output is
used as a decision score, and separation between groups does not imply that its
numerical value equals a downstream failure probability.


\begin{figure}[!htbp]
\centering
\includegraphics[width=\linewidth]{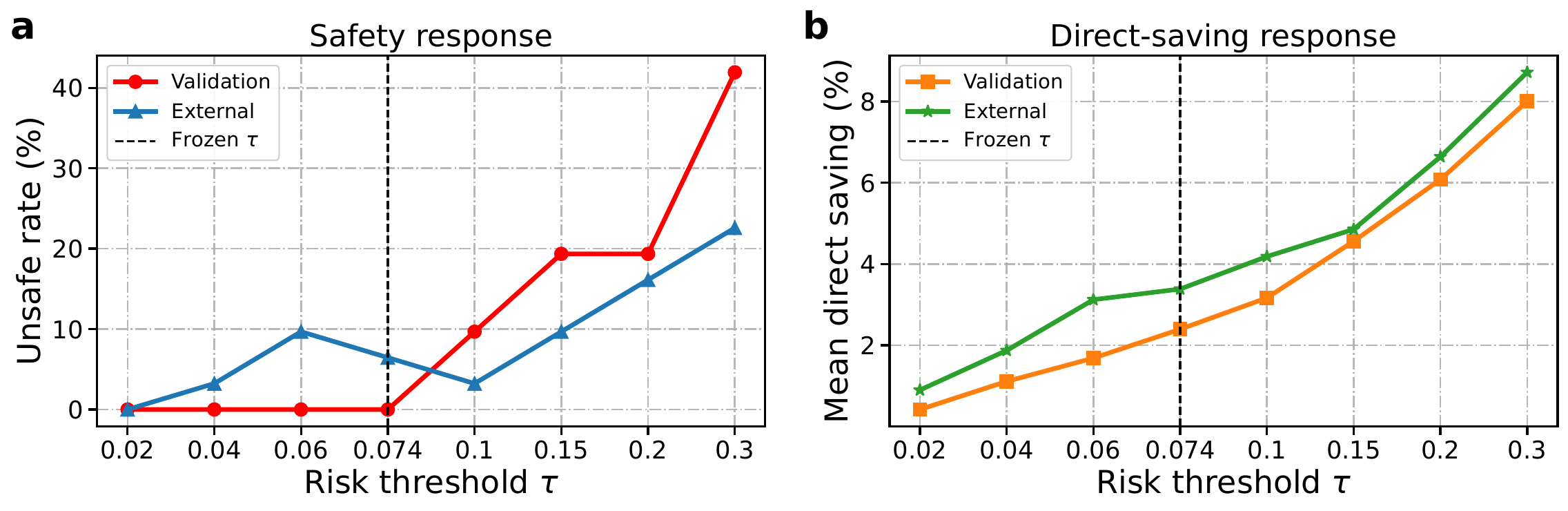}
\caption{Risk-gate operating curve. Left: empirical unsafe rate. Right: mean direct removal. The dashed line marks the frozen deployment threshold.}
\label{fig:risk_threshold_appendix}
\end{figure}

Figure~\ref{fig:risk_threshold_appendix} visualizes the operating behavior of
the risk gate as its threshold changes. The left panel tracks empirical unsafe
rate, while the right panel tracks mean direct removal, exposing the trade-off
between conservative selection and more aggressive compression. The dashed
line marks the threshold selected on validation and subsequently frozen for
deployment. Thus, the online threshold is chosen as a policy operating point,
rather than being identified with the offline harm threshold
$\delta_H$.


\begin{figure}[!htbp]
\centering
\includegraphics[width=0.52\linewidth]{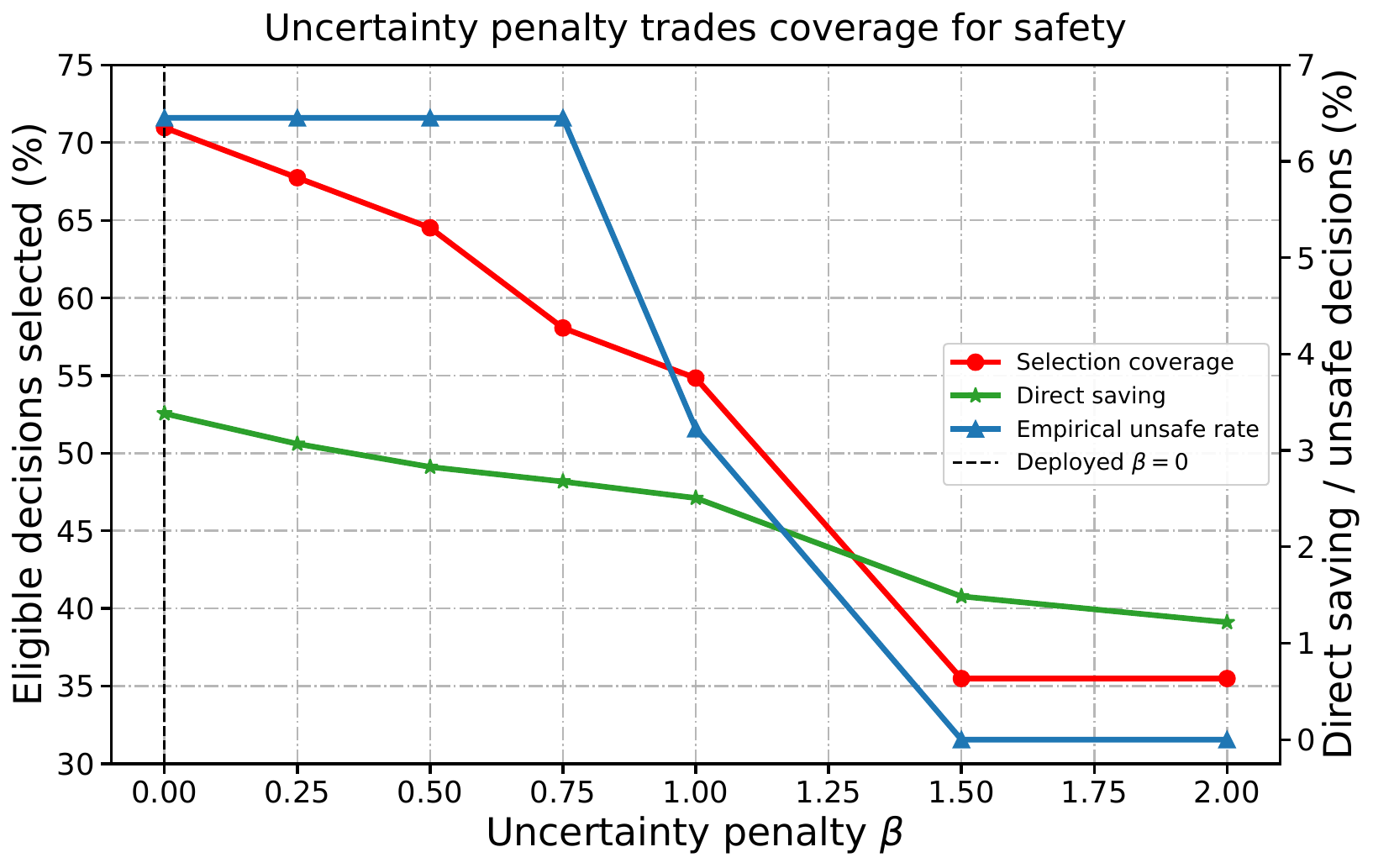}
\caption{Selection coverage, direct saving, and empirical unsafe rate across the uncertainty coefficient $\beta$. The deployed setting is $\beta=0$.}
\label{fig:uncertainty}
\end{figure}

Figure~\ref{fig:uncertainty} reports sensitivity to the ensemble uncertainty
coefficient $\beta$. Increasing $\beta$ changes the deployment score from the
ensemble mean toward a more uncertainty-penalized criterion, which can alter
selection coverage, direct removal, and empirical unsafe rate. The validation
search selects $\beta=0$ for the deployed configuration, so the final policy
uses the ensemble-mean unsafe score without an additional standard-deviation
penalty. This analysis shows that the uncertainty term was explicitly tested
rather than implicitly fixed.


\begin{figure}[!htbp]
\centering
\includegraphics[width=\linewidth]{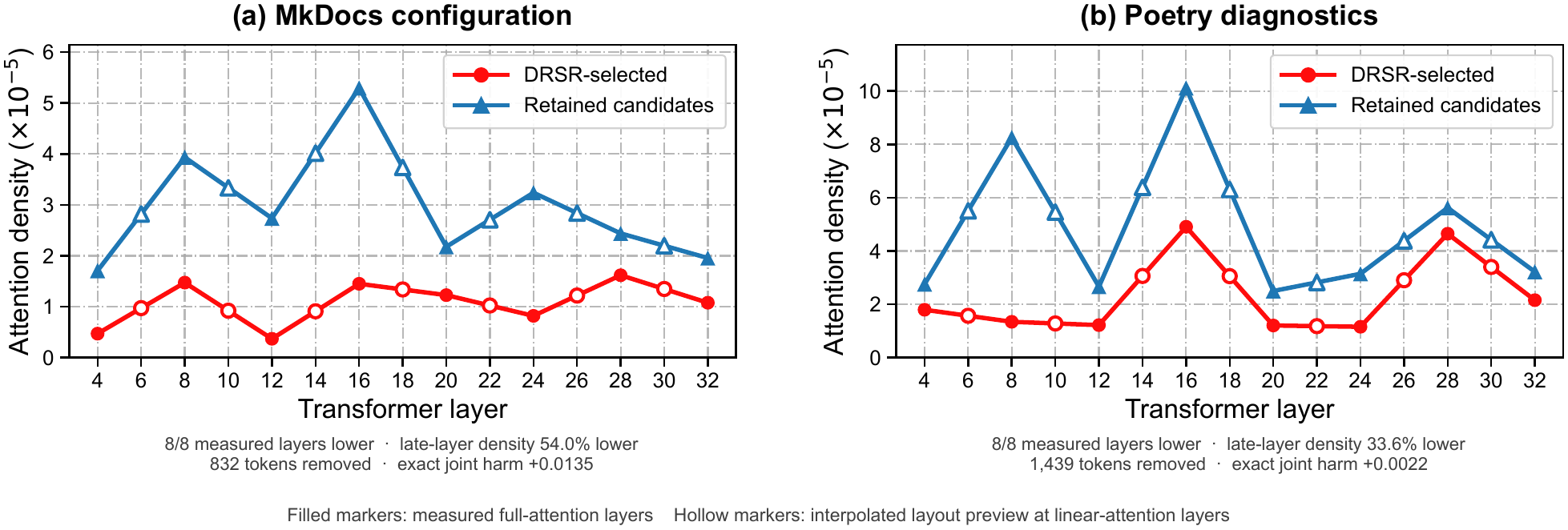}
\caption{Attention density for selected and retained candidates in two examples. Filled markers are measured full-attention layers; hollow markers are interpolated layout points. Lower attention alone is not a deletion-safety test.}
\label{fig:attention_cases}
\end{figure}

Figure~\ref{fig:attention_cases} provides two qualitative examples of attention
density across representation depth for candidates that are selected for
deletion and candidates that are retained. Filled markers correspond to
measured full-attention layers, whereas hollow markers are interpolation points
used only for visualization. The examples illustrate why attention is treated
as one relational feature rather than a standalone pruning criterion: the
deletion decision is produced by the learned set-risk model together with
state, retained-context, and interaction information, not by applying a fixed
low-attention threshold.

\FloatBarrier
\section{Optimization and Metric Definitions}
\label{app:optimization}

\subsection{Harm Standardization and Ranking}
Let $\mu_H$ and $\sigma_H$ be computed on training labels only. Regression uses
\begin{equation}
 \widetilde H_t(A)=\frac{H_t(A)-\mu_H}{\max(\sigma_H,0.02)}.
 \label{eq:standardize}
\end{equation}
Predictions are restored to the original harm scale for interpretation. Ranking pairs have the same source trajectory and deletion cardinality. For examples $a=(t,A)$ and $b=(t',B)$, define $\Delta_{ab}=H_t(A)-H_{t'}(B)$. Pairs with $|\Delta_{ab}|<10^{-5}$ are omitted. With
\begin{equation}
 \omega_{ab}=0.1+0.9\min\left(\frac{|\Delta_{ab}|}{0.02},1\right),
 \label{eq:rankweight}
\end{equation}
training minimizes the weighted pairwise loss
\begin{equation}
 \mathcal L_{\mathrm{rank}}
 =\mathbb E_{a,b}\left[
  \omega_{ab}\operatorname{softplus}\left(
  -\operatorname{sign}(\Delta_{ab})(\widehat H_a-\widehat H_b)
  \right)\right].
 \label{eq:ranking}
\end{equation}
Trajectory-grouped training pairs and within-state evaluation pairs are different grouping rules. Both preserve the intended source dependence; their metrics should not be silently interchanged.

\subsection{Offline Prediction Metrics}
For a set of evaluated pairs $\mathcal P$ satisfying the specified harm margin, pair accuracy is the fraction with correctly ordered predicted harm. Tied predictions need an explicit scoring convention. Within-state Spearman correlation compares the predicted and measured set orders at a common decision state and cardinality. A pooled accuracy and a mean of trajectory-level accuracies generally differ, so an empirical report must identify which aggregation was used.

The unsafe target is $u_t(A)$ from Eq.~\eqref{eq:label}. Brier score averages $(\widehat p_t(A)-u_t(A))^2$, while AUROC measures the ability to rank unsafe events above safe events. Neither is a task reward metric. Multiple sets from one trajectory are dependent observations; uncertainty estimates require grouping at the trajectory level rather than treating every deletion row as independent.

\subsection{Direct and Complete-Task Token Accounting}
For task $j$ and requests $t$, direct deletion is
\begin{equation}
 D_{\mathrm{direct},j}=\sum_t\left[
 \operatorname{Tok}(C_{j,t})-
 \operatorname{Tok}(D_{A_{j,t}}(C_{j,t}))\right].
 \label{eq:direct}
\end{equation}
Complete-task processed tokens are instead $T_j=\sum_r(I_{j,r}+O_{j,r})$. Cache-read and cache-creation counts, when reported as input components, are already included in $I_{j,r}$ and are not added again. For benchmark means,
\begin{equation}
 \Delta_T(\%)=100\left(\frac{\bar T_{\mathrm{DRSR}}}{\bar T_{\mathrm{Base}}}-1\right).
 \label{eq:tokenchange}
\end{equation}
A negative token change denotes a decrease. A positive ``reduction'' percentage would use the opposite sign. Monetary cost cannot be inferred from this unweighted token total without the applicable prices and input/output/cache breakdown.

Changes in action paths, retries, termination, and request counts can separate direct removal from the observed difference between two full rollouts. One may define a residual $\Delta T_{\mathrm{rollout}}$ so that
\begin{equation}
 T_{\mathrm{Base}}-T_{\mathrm{DRSR}}
 =D_{\mathrm{direct}}+\Delta T_{\mathrm{rollout}},
 \label{eq:accounting}
\end{equation}
but this is an accounting decomposition, not causal identification of the residual. Local representation computation is a separate workload and is not automatically captured by API token totals.

\section{Comparator Token Accounting}
\label{app:comparator_tokens}

Table~\ref{tab:comparator_tokens} reports the Eval40 token quantities used in
Table~\ref{tab:comparison}. For methods with an auxiliary summarization model,
its reported input and output tokens are included in the total. Totals are
shown in millions to keep the appendix compact; the per-task column uses the
same accounting rule as the main comparison.

\begin{table}[t]
\centering
\caption{Eval40 token accounting for comparator methods. Totals cover 40 tasks.}
\label{tab:comparator_tokens}
\tableformat
\fontsize{9.1}{10.8}\selectfont
\setlength{\tabcolsep}{8pt}
\renewcommand{\arraystretch}{1.04}
\begin{tabular*}{0.78\linewidth}{@{\extracolsep{\fill}}lrr@{}}
\toprule[0.9pt]
\textbf{Method} & \textbf{Total (M)} & \textbf{Per task (M)} \\
\midrule
DeepSeek-V4-Flash          & 75.50 & 1.887 \\
Sliding Window ($K=5$)     & 118.08 & 2.952 \\
Sliding Window ($K=10$)    & 79.39 & 1.985 \\
Sliding Window ($K=20$)    & 80.82 & 2.020 \\
Periodic Summary ($n=3$)   & 88.57 & 2.214 \\
Periodic Summary ($n=5$)   & 94.85 & 2.371 \\
\midrule
PACE                        & 44.12 & 1.103 \\
LLMLingua-2                 & 96.18 & 2.405 \\
SelfCompact                 & 62.70 & 1.567 \\
ACON-Core                   & 56.13 & 1.403 \\
Self-GC                     & 66.21 & 1.655 \\
LRE                         & 91.64 & 2.291 \\
CoMem                       & \best{30.64} & \best{0.766} \\
SAM                         & 53.85 & 1.346 \\
SWE-Pruner                  & 60.07 & 1.502 \\
Sculptor                    & 51.75 & 1.294 \\
ACM                         & 94.56 & 2.364 \\
\midrule
\textbf{DRSR}               & \textbf{48.43} & 1.211 \\
\bottomrule[0.9pt]
\end{tabular*}
\end{table}

\section{Detailed Task and Token Summaries}
\label{app:domain_results}

This section provides the domain-level quantities underlying the Eval40
ablation and the Full260 token summaries. To improve readability, token
magnitudes are reported in millions (M) or thousands (K) rather than as
long raw counts. This changes only presentation, not the underlying values.

\subsection{Domain-Level Ablations}

Table~\ref{tab:ablation_domains} preserves the four-domain Eval40 ablation
results. Each domain has equal weight in the Eval40 summary; for example,
the full DRSR reward is
\begin{equation}
(0.856+0.788+0.700+0.833)/4=0.794.
\label{eq:eval40mean}
\end{equation}

\begin{table}[t]
\centering
\caption{
Per-domain Eval40 ablations.
$T$ is mean total model tokens per task in millions;
Direct is directly removed tokens per task in thousands.
}
\label{tab:ablation_domains}
\tableformat
\fontsize{9.0}{10.7}\selectfont
\setlength{\tabcolsep}{8pt}
\renewcommand{\arraystretch}{1.04}
\begin{tabular*}{0.80\linewidth}{@{\extracolsep{\fill}}lrrr@{}}
\toprule[0.9pt]
\textbf{Domain} & \textbf{Reward} & \textbf{$T$ (M)} & \textbf{Direct (K)} \\
\midrule
\multicolumn{4}{l}{\textit{DRSR}} \\[-1pt]
Code     & 0.856 & 1.216 & 46.6 \\
Office   & 0.788 & 0.724 &  6.8 \\
Security & 0.700 & 1.616 & 37.6 \\
Web      & 0.833 & 1.288 &  9.7 \\
\addlinespace[2pt]
\multicolumn{4}{l}{\textit{w/o retained}} \\[-1pt]
Code     & 0.722 & 0.716 & 22.7 \\
Office   & 0.779 & 0.266 &  0.0 \\
Security & 0.698 & 1.254 & 17.7 \\
Web      & 0.800 & 1.731 & 10.1 \\
\addlinespace[2pt]
\multicolumn{4}{l}{\textit{w/o pair pooling}} \\[-1pt]
Code     & 0.756 & 0.567 & 40.3 \\
Office   & 0.678 & 0.606 &  2.3 \\
Security & 0.796 & 1.094 & 52.1 \\
Web      & 0.796 & 2.535 & 34.1 \\
\addlinespace[2pt]
\multicolumn{4}{l}{\textit{w/o abstention}} \\[-1pt]
Code     & 0.833 & 1.205 & 46.9 \\
Office   & 0.801 & 0.447 &  1.6 \\
Security & 0.573 & 0.963 & 45.7 \\
Web      & 0.800 & 1.553 & 26.3 \\
\bottomrule[0.9pt]
\end{tabular*}
\end{table}

The ablations are visibly domain dependent. Removing retained-context features
reduces direct deletion in several domains but still lowers overall reward;
removing pair pooling produces the largest direct deletion in Security; and
removing abstention is particularly harmful in Security. These domain-level
patterns are consistent with the main-text conclusion that the three components
affect different parts of the pruning decision rather than acting as redundant
copies of the same mechanism.

\subsection{Token Components}
\label{app:token_results}

Table~\ref{tab:tokens} gives the equal-domain token decomposition used by the
Eval40 ablation summary. Input and cache quantities are shown in millions of
tokens per task, while output is shown in thousands because it is an order of
magnitude smaller. Cache read and cache creation are components of input and
are therefore not added to the total a second time.

\begin{table}[t]
\centering
\caption{
Equal-domain token decomposition on Eval40.
Input, cache, and total columns are in millions of tokens per task;
output is in thousands.
}
\label{tab:tokens}
\tableformat
\fontsize{9.0}{10.7}\selectfont
\setlength{\tabcolsep}{6.5pt}
\renewcommand{\arraystretch}{1.04}
\begin{tabular*}{0.92\linewidth}{@{\extracolsep{\fill}}lrrrrr@{}}
\toprule[0.9pt]
\textbf{Variant}
& \textbf{Input (M)}
& \textbf{Output (K)}
& \textbf{Cache read (M)}
& \textbf{Cache create (M)}
& \textbf{Total (M)} \\
\midrule
DRSR            & 1.188 & 22.7 & 1.067 & 0.121 & 1.211 \\
w/o retained    & 0.970 & 21.3 & 0.873 & 0.097 & 0.992 \\
w/o pair pooling& 1.178 & 22.2 & 1.052 & 0.126 & 1.201 \\
w/o abstention  & 1.018 & 23.7 & 0.872 & 0.147 & 1.042 \\
\bottomrule[0.9pt]
\end{tabular*}
\end{table}

The decomposition shows that the observed token differences are driven mainly
by input-side processing rather than output length. This is expected for
history pruning: changing what remains in the active context primarily changes
how much prior context is repeatedly processed. The online controller itself
uses two local representation forwards at eligible decisions; that local
workload is separate from the API token totals reported here.

For completeness, Table~\ref{tab:token_domains} retains the per-domain token
components in the same compact units.

\begin{table}[t]
\centering
\caption{
Per-domain token components for Eval40 ablations.
Input and cache columns are in millions of tokens per task;
output is in thousands.
}
\label{tab:token_domains}
\tableformat
\fontsize{8.9}{10.5}\selectfont
\setlength{\tabcolsep}{6pt}
\renewcommand{\arraystretch}{1.03}
\begin{tabular*}{0.90\linewidth}{@{\extracolsep{\fill}}lrrrr@{}}
\toprule[0.9pt]
\textbf{Domain}
& \textbf{Input (M)}
& \textbf{Output (K)}
& \textbf{Cache read (M)}
& \textbf{Cache create (M)} \\
\midrule
\multicolumn{5}{l}{\textit{DRSR}} \\[-1pt]
Code     & 1.205 & 11.2 & 1.052 & 0.153 \\
Office   & 0.702 & 22.0 & 0.642 & 0.060 \\
Security & 1.589 & 26.4 & 1.429 & 0.161 \\
Web      & 1.256 & 31.0 & 1.146 & 0.110 \\
\addlinespace[2pt]
\multicolumn{5}{l}{\textit{w/o retained}} \\[-1pt]
Code     & 0.706 & 10.1 & 0.605 & 0.101 \\
Office   & 0.247 & 19.4 & 0.196 & 0.050 \\
Security & 1.232 & 21.8 & 1.095 & 0.137 \\
Web      & 1.697 & 33.9 & 1.596 & 0.100 \\
\addlinespace[2pt]
\multicolumn{5}{l}{\textit{w/o pair pooling}} \\[-1pt]
Code     & 0.558 &  9.0 & 0.463 & 0.095 \\
Office   & 0.583 & 23.6 & 0.519 & 0.064 \\
Security & 1.072 & 21.7 & 0.881 & 0.191 \\
Web      & 2.501 & 34.5 & 2.345 & 0.156 \\
\addlinespace[2pt]
\multicolumn{5}{l}{\textit{w/o abstention}} \\[-1pt]
Code     & 1.194 & 11.4 & 1.014 & 0.179 \\
Office   & 0.425 & 22.1 & 0.349 & 0.076 \\
Security & 0.940 & 23.2 & 0.769 & 0.171 \\
Web      & 1.515 & 37.9 & 1.355 & 0.160 \\
\bottomrule[0.9pt]
\end{tabular*}
\end{table}

\subsection{Full260 Token Accounting}

Table~\ref{tab:full_tokens} reports the Full260 reward and token quantities in
the same compact notation. $T$ is the complete-task model-token mean per task,
whereas Direct measures only the tokens removed from the rewritten request at
the pruning decision. The two quantities should not be conflated because
different pruning decisions can also change later action paths, request counts,
termination, and subsequent context growth.

\begin{table}[t]
\centering
\caption{
Full260 reward and token accounting by domain.
Token columns are per-task means in millions;
Direct is per-task direct removal in thousands.
}
\label{tab:full_tokens}
\tableformat
\fontsize{8.9}{10.6}\selectfont
\setlength{\tabcolsep}{5.2pt}
\renewcommand{\arraystretch}{1.04}
\begin{tabular*}{0.96\linewidth}{@{\extracolsep{\fill}}lrrrrrrr@{}}
\toprule[0.9pt]
\textbf{Domain}
& \textbf{$N$}
& \textbf{Base $R$}
& \textbf{DRSR $R$}
& \textbf{Base $T$ (M)}
& \textbf{DRSR $T$ (M)}
& \textbf{$\Delta T$}
& \textbf{Direct (K)} \\
\midrule
Code     & 80 & 0.770 & 0.858 & 1.370 & 1.235 & $-9.91\%$  & 46.9 \\
Office   & 50 & 0.818 & 0.824 & 1.198 & 0.997 & $-16.82\%$ &  6.8 \\
Security & 60 & 0.478 & 0.756 & 5.977 & 4.569 & $-23.55\%$ & 31.0 \\
Web      & 70 & 0.721 & 0.760 & 2.429 & 1.858 & $-23.52\%$ & 38.2 \\
\midrule
\textbf{Overall}
& \textbf{260}
& \textbf{0.699}
& \textbf{0.802}
& \textbf{2.685}
& \textbf{2.126}
& \textbf{$-20.82\%$}
& \textbf{33.2} \\
\bottomrule[0.9pt]
\end{tabular*}
\end{table}

The Full260 table makes the domain dependence of compression explicit.
Security has by far the largest baseline context and also shows the largest
absolute reward improvement, while Office changes only modestly. DRSR reduces
processed tokens in all four domains, but the amount of direct same-request
removal is not identical to the complete-rollout token difference.

\FloatBarrier

\section{Deployment and Supervision Details}
\label{app:deployment_details}

\subsection{Training and Deployment Information Boundary}
Offline supervision may use quantities that are unavailable when the online pruning decision is made. Table~\ref{tab:information_boundary} makes this separation explicit. The realized next output $y_t$ and exact deletion harm are used only to construct supervision. The deployed scorer receives only the current causal prefix, candidate features, and the proposed deletion mask.

\begin{table}[!htbp]
\centering
\caption{Information boundary for DRSR. ``Label only'' means that a quantity may construct supervision but is not an online feature.}
\label{tab:information_boundary}
\tableformat
\begin{tabularx}{0.96\linewidth}{@{}LYYY@{}}
\toprule
\textbf{Quantity} & \textbf{Offline training} & \textbf{Online selection} & \textbf{Post-hoc audit} \\
\midrule
Current query and observed history & Yes & Yes & Yes \\
Pre-action hidden states / attention & Yes & Yes & Yes \\
Candidate mask and token saving & Yes & Yes & Yes \\
Realized next output $y_t$ & Label only & No & Yes \\
Exact joint deletion harm $H_t(A)$ & Supervision & No & Optional remeasurement \\
Task reward & Not a feature & No & System evaluation \\
Counterfactual deletion forward & Label generation & 0 & Optional audit \\
\bottomrule
\end{tabularx}
\end{table}

\subsection{Exact Online Risk Aggregation and Feasible Set}
For ensemble member $s$, let $\widehat H_t^{(s)}(A)$ and $\widehat p_t^{(s)}(A)$ denote the predicted harm and unsafe score. DRSR aggregates the three members as
\begin{equation}
 \bar H_t(A)=\frac{1}{3}\sum_s\widehat H_t^{(s)}(A),\qquad
 U_t(A)=\bar p_t(A)+\beta\,\Std_s\!\left[\widehat p_t^{(s)}(A)\right].
 \label{eq:risk_score}
\end{equation}
The deployed setting uses $\beta=0$, so $U_t(A)$ is the ensemble-mean unsafe score. The exact feasible set is
\begin{equation}
\begin{aligned}
\mathcal F_t=\{A\subseteq E_t:\;&1\le |A|\le K,\quad
0<w_t(A)\le q_{\max}\operatorname{Tok}(C_t),\\
&A\cap\mathrm{Recent2}=\varnothing,\quad
U_t(A)\le\tau,\quad
\operatorname{ProtocolSafe}(D_A(C_t))\}.
\end{aligned}
\label{eq:feasible}
\end{equation}
The main-text set $\mathcal A_t$ contains the same structural constraints without the risk condition. Equation~\eqref{eq:selection_objective} then chooses the largest-saving member that also satisfies $U_t(A)\le\tau$. If several sets remove the same number of tokens, the policy prefers lower unsafe score and then lower predicted harm.

With at most four candidates and $K=3$, the number of nonempty subsets considered online is
\begin{equation}
 \sum_{k=1}^{3}\binom{4}{k}=14.
 \label{eq:subset_count}
\end{equation}
All candidate subsets are batch-scored by the lightweight network. No online counterfactual deletion forward is executed.

\subsection{Online Computational Cost}
\label{app:online_cost}

The online controller does not evaluate deletion harm by rerunning the
language model on each candidate rewrite. At an eligible decision, the
implementation performs two local representation passes over the intact
causal prefix: one pass extracts attention and hidden-state relations, and
the second provides the projected semantic representation used by the
feature pipeline. Candidate-specific features are then assembled from these
intact-prefix representations.

With $n\le4$ candidates and $K=3$, at most
\[
\sum_{k=1}^{3}\binom{n}{k}\le14
\]
candidate sets are formed. Their 451-dimensional set representations are
scored together by the lightweight scorer ensemble. The number of online
candidate-specific counterfactual deletion forwards is therefore exactly
zero. Only the single deletion set selected by the policy is applied to the
request that is passed to the acting model.

This distinction is important for both meaning and cost. The offline label
$H_t(A)$ compares the likelihood of one fixed recorded output under intact
and edited contexts. Generating separate online continuations for each
candidate would not recover that fixed-target quantity, and it would add a
large inference cost to the pruning procedure itself.

\subsection{Protocol-Preserving Deletion}
The implementation enforces the following invariants at every eligible decision: (i) the two latest complete Steps are protected; (ii) a tool call and its matched result are atomic; (iii) no rewrite may create dangling tool identifiers or an empty assistant message; (iv) actual removal on the full serialized request is at most 20\%; (v) only subsets passing the frozen risk gate are eligible; (vi) if no subset is feasible, the complete request is retained; (vii) alignment or protocol-validation failure also returns the request unchanged; and (viii) online selection never reads a future target and never executes a counterfactual deletion forward. These checks are logically separate from the learned risk score.

\subsection{Exact Set-Level Supervision}
\label{app:label_construction}
For a selected decision state, candidate construction is frozen before any deletion outcome is observed. Three candidates generate all seven nonempty subsets, while four candidates generate all fifteen. Every subset is jointly deleted from the original intact context, the request is reserialized, and the same recorded next output is teacher-forced under the full and deleted contexts. Joint labels are therefore measurements of the actual joint rewrite, not sums of singleton labels.

All states and deletion sets originating from one source trajectory stay in the same 80:10:10 train/validation/test partition. This prevents deletion rows from a common trajectory from crossing splits. The supervision corpus contains 566 external trajectories and 1,698 target states, with approximately 24,122 set-level labels. The cardinality breakdown is 6,623 singletons, 9,682 pairs, 6,287 triples, and 1,530 quadruples.

\begin{table}[!htbp]
\centering
\caption{Set-level supervision by deletion cardinality.}
\label{tab:label_cardinality}
\tableformat
\begin{tabularx}{0.70\linewidth}{@{}LYY@{}}
\toprule
\textbf{Deletion size} & \textbf{Labelled sets} & \textbf{Mean interaction} \\
\midrule
1 & 6,623 & 0 \\
2 & 9,682 & 0.008 \\
3 & 6,287 & 0.016 \\
4 & 1,530 & 0.027 \\
\bottomrule
\end{tabularx}
\tablenote{The interaction statistic is the source-reported summary used in the joint-deletion analysis. No additive approximation is used to construct the labels.}
\end{table}

\subsection{Validation Selection and Frozen Deployment Policy}
\label{app:calibration}
Validation searches the maximum online deletion cardinality, an optional ensemble uncertainty penalty, and the deployment threshold. A candidate policy is admissible only when its empirical validation unsafe rate is at most 0.10 and its mean selected-set exact harm is at most 0.02. Among admissible policies, validation maximizes the mean fraction of the full request removed, then prefers lower unsafe rate and lower harm. Test labels and WorkBuddyBench outcomes are not used for this selection.

\begin{table}[!htbp]
\centering
\caption{Validation search space and frozen deployment values.}
\label{tab:calibration}
\tableformat
\begin{tabularx}{0.86\linewidth}{@{}Lcc@{}}
\toprule
\textbf{Item} & \textbf{Validation search / rule} & \textbf{Frozen value} \\
\midrule
Maximum cardinality $K$ & $\{2,3\}$ & 3 \\
Uncertainty coefficient $\beta$ & $\{0,0.25,0.5,1,2\}$ & 0 \\
Risk threshold $\tau$ & observed validation scores & 0.0740822119017442 \\
Empirical unsafe-rate constraint & $\le 0.100$ & enforced in selection \\
Mean selected exact-harm constraint & $\le 0.020$ & enforced in selection \\
Full-request removal cap & fixed & 20\% \\
Protected history & fixed & Recent2 \\
Fallback & fixed & strict abstention \\
\bottomrule
\end{tabularx}
\end{table}

The harm-label threshold $\delta_H=0.02$ and the deployment threshold $\tau=0.0740822119017442$ act on different quantities. The first thresholds measured teacher-forced NLL change to create an offline binary target. The second thresholds the scorer's predicted unsafe score during online selection. Their numerical values therefore need not match.

\FloatBarrier

\Needspace{550pt}
\section{Eval40 Task Manifest}
\label{app:eval40_manifest}

Eval40 contains ten tasks from each of Code, Office, Security, and Web. The task identifiers
below are the frozen manifest used by the reported context-management comparison. Listing
the identifiers makes the 40-task comparison reproducible and distinguishes it from other
40-task pilots.

\begingroup
\fontsize{8.5}{10.2}\selectfont
\setlength{\LTpre}{4pt}
\setlength{\LTpost}{4pt}
\begin{longtable}{@{}p{0.10\linewidth}p{0.08\linewidth}p{0.74\linewidth}@{}}
\toprule
\textbf{Domain} & \textbf{\#} & \textbf{Task ID} \\
\midrule
\endfirsthead
\toprule
\textbf{Domain} & \textbf{\#} & \textbf{Task ID} \\
\midrule
\endhead
Code & 1 & \texttt{api\_contract-hard-markup\_errors} \\
Code & 2 & \texttt{api\_contract-hard-openapi\_params} \\
Code & 3 & \texttt{api\_contract-hard-token\_errors} \\
Code & 4 & \texttt{api\_contract-hard-validation\_errors} \\
Code & 5 & \texttt{bug\_fix-easy-a\_crash\_in\_local} \\
Code & 6 & \texttt{bug\_fix-easy-filtered\_relation\_queryset\_arg} \\
Code & 7 & \texttt{bug\_fix-easy-invalid\_filterwarnings\_regex\_error} \\
Code & 8 & \texttt{bug\_fix-medium-error\_key\_uses\_data\_key} \\
Code & 9 & \texttt{bug\_fix-medium-errors\_from\_earlier\_indices} \\
Code & 10 & \texttt{bug\_fix-medium-incorrect\_linenos\_on\_fstring} \\
\midrule
Office & 11 & \texttt{analyst-forecast-extract-L3-018} \\
Office & 12 & \texttt{api-usage-explain-cli-l3-001} \\
Office & 13 & \texttt{board-material-update-timeline-excel} \\
Office & 14 & \texttt{calendar-dida-sync-state} \\
Office & 15 & \texttt{channel-period-compare-L4-017} \\
Office & 16 & \texttt{cloudagent-sdk-doc-validation-report} \\
Office & 17 & \texttt{contract-extract-L3-014} \\
Office & 18 & \texttt{cross-week-dashboard-migration} \\
Office & 19 & \texttt{crypto-backtest-chain-L4-002} \\
Office & 20 & \texttt{daily-creation-checkpoint-recovery} \\
\midrule
Security & 21 & \texttt{agent-to-agent-injection-hard-multistep} \\
Security & 22 & \texttt{apt-multi-source-correlation-hard-multistep} \\
Security & 23 & \texttt{bb-bin-dns-parse-010} \\
Security & 24 & \texttt{bb-bin-firmware-audit-007} \\
Security & 25 & \texttt{bb-bin-format-log-004} \\
Security & 26 & \texttt{bb-bin-int-length-005} \\
Security & 27 & \texttt{bb-bin-ipc-cache-001} \\
Security & 28 & \texttt{bb-bin-media-parse-008} \\
Security & 29 & \texttt{bb-bin-oob-read-003} \\
Security & 30 & \texttt{bb-bin-parse-crash-006} \\
\midrule
Web & 31 & \texttt{animated-explainer-L3-028} \\
Web & 32 & \texttt{atmosphere-game-L4-035} \\
Web & 33 & \texttt{blog-editor-draft-recovery-L4-059} \\
Web & 34 & \texttt{browser-clipper-extension-L4-005} \\
Web & 35 & \texttt{canvas-webgl-scene-L4-026} \\
Web & 36 & \texttt{chart-generation-L2-025} \\
Web & 37 & \texttt{checkout-incident-analysis-L4-049} \\
Web & 38 & \texttt{city-article-theme-variants-L4-066} \\
Web & 39 & \texttt{claims-drawer-state-review-report-L3-071} \\
Web & 40 & \texttt{cohort-retention-dashboard-L4-054} \\
\bottomrule
\end{longtable}
\endgroup
\end{document}